\documentclass[dvipsnames]{article} 
\usepackage{iclr2025_conference,times}
\usepackage{microtype}
\usepackage{graphicx}
\usepackage{multirow}
\usepackage{xcolor, colortbl}  
\usepackage{subfig}
\usepackage{booktabs,array} 

\usepackage{wrapfig}

\usepackage{pifont}

\usepackage{amsmath}
\usepackage{amssymb}
\usepackage{mathtools}
\usepackage{amsthm}

\theoremstyle{plain}

\theoremstyle{definition}

\theoremstyle{remark}

\usepackage[textsize=tiny]{todonotes}
\usepackage{xspace}

\usepackage{url}

\definecolor{Gray}{gray}{0.85}
\definecolor{Yellow}{rgb}{0.5,0.5,0.5}

\usepackage{comment}
\usepackage{epsfig}
\usepackage[backref=page,hidelinks]{hyperref}
\usepackage{mathrsfs,mdwlist,enumitem}
\usepackage[group-separator={,}]{siunitx}

\newcommand{\permute}{\textsc{Permute}\xspace}
\newcommand{\git}{\textsc{Git-Rebasin}\xspace}
\newcommand{\wa}{\textsc{WAvg.}\xspace}
\newcommand{\name}{\textsc{GnnMerge}\xspace}
\newcommand{\surgery}{\textsc{Surgery}\xspace}
\newcommand{\zipit}{\textsc{ZipIt!}\xspace}

\newcommand{\gnns}{\textsc{Gnn}s\xspace}
\newcommand{\gnn}{\textsc{Gnn}\xspace}
\newcommand{\node}{\textsc{NodeFormer}\xspace}
\newcommand{\gat}{\textsc{Gat}\xspace}
\newcommand{\sage}{\textsc{GraphSage}\xspace}
\newcommand{\gcn}{\textsc{Gcn}\xspace}
\newcommand{\gin}{\textsc{Gin}\xspace}
\newcommand{\mlp}{\textsc{MLP}\xspace}
\newcommand{\mpnn}{\textsc{Mpnn}\xspace}
\newcommand{\CG}{\mathcal{G}\xspace}

\newcommand{\CY}{\mathcal{Y}\xspace}

\newcommand{\CN}{\mathcal{N}\xspace}
\newcommand{\CT}{\mathcal{T}\xspace}

\newcommand{\CV}{\mathcal{V}\xspace}
\newcommand{\CE}{\mathcal{E}\xspace}

\newcommand{\X}{\boldsymbol{X}\xspace}

\newcommand{\CZ}{\mathbf{Z}\xspace}
\newcommand{\CS}{\mathcal{S}\xspace}

\newcommand{\cW}{\mathbf{W}\xspace}
\newcommand{\cS}{\mathbf{S}\xspace}
\newcommand{\cG}{\mathbf{G}\xspace}
\newcommand{\CL}{\mathcal{L}\xspace}
\newcommand{\cx}{\mathbf{x}\xspace}

\newcommand{\cm}{\mathbf{m}\xspace}
\newcommand{\cg}{\mathbf{g}\xspace}

\newcommand{\ch}{\mathbf{h}\xspace}

\newcommand{\cz}{\mathbf{z}\xspace}

\newcommand{\relu}{\text{ReLU}}

\setlist{nolistsep,leftmargin=*}

\newtheorem{defn}{\textbf{Definition}}

\newtheorem{prob}{\textbf{Problem}}

\definecolor{babypink}{rgb}{0.96, 0.76, 0.76}
\definecolor{top1}{HTML}{a5dc82}
\definecolor{top2}{HTML}{dff3d9}

\definecolor{c1}{HTML}{d5e8d4}
\definecolor{c1_1}{HTML}{82b366}
\definecolor{c2}{HTML}{ffe6cc}
\definecolor{c2_1}{HTML}{d79b01}
\definecolor{c3}{HTML}{dae8fc}
\definecolor{c3_1}{HTML}{6c8ebf}
\definecolor{text_grey}{HTML}{5e5e5e}
\hypersetup{hidelinks}
\renewcommand{\backref}[1]{}
\renewcommand{\backrefalt}[4]{%
    \ifcase #1%
    \or%
    (Cited on p.~\bfseries #2 \(\boldsymbol{\hookleftarrow}\))%
    \else%
    (Cited on pp.~\bfseries #2 \(\boldsymbol{\hookleftarrow}\))%
    \fi
}

\usepackage[capitalize,noabbrev]{cleveref}

\title{\name: Merging of \gnn Models Without Accessing Training Data}

\author{Vipul Garg, Ishita Thakre \& Sayan Ranu \\
Department of Computer Science\\
Indian Institute of Technology Delhi, New Delhi, 110016, India\\
\texttt{\{cs5200450,cs5200445,sayanranu\}@cse.iitd.ac.in} \\
}

\iclrfinalcopy 
\begin{document}

\maketitle

\begin{abstract}
Model merging has gained prominence in machine learning as a method to integrate multiple trained models into a single model without accessing the original training data. While existing approaches have demonstrated success in domains such as computer vision and NLP, their application to Graph Neural Networks (GNNs) remains unexplored. These methods often rely on the assumption of shared initialization, which is seldom applicable to GNNs. In this work, we undertake the first benchmarking study of model merging algorithms for GNNs, revealing their limited effectiveness in this context. To address these challenges, we propose \name, which utilizes a task-agnostic node embedding alignment strategy to merge \gnns. Furthermore, we establish that under a mild relaxation, the proposed optimization objective admits direct analytical solutions for widely used GNN architectures, which significantly enhances its computational efficiency. Empirical evaluations across diverse datasets, tasks, and architectures establish \name to be up to $24\%$ more accurate than existing methods while delivering over $2$ orders of magnitude speed-up compared to training from scratch.
\end{abstract}

\vspace{-0.2in}
\section{Introduction}
\label{sec:intro}
\vspace{-0.15in}
Given two neural models, can we \textit{merge} them into a single model integrating the capabilities of both, without accessing the original training data? This is the core question driving the emergent field of \textit{model merging}~\cite{stoica2024zipit,ainsworth2023gitrebasin, yang2024adamerging,ilharco2023editingtaskarithmetic,emr-merging-DBLP:journals/corr/abs-2405-17461,lu2024twinmerging}. Model merging addresses key challenges in dynamic machine learning environments where  retraining-from-scratch is impractical or impossible. For instance, the introduction of training data annotated with new class labels—such as novel research areas in citation networks or new product types in e-commerce— necessitates full retraining after incorporating the new training data. A more efficient alternative would be to train a new model \textit{exclusively} on the new data and then merge it with the existing model, and thereby eliminating the need for full retraining. Similarly, one may wish to merge two models trained on the same dataset but for different tasks into a single multi-task model. In privacy-sensitive settings, organizations may wish to combine independently trained models without sharing raw data, avoiding privacy breaches or exposing proprietary information. 
\vspace{-0.1in}
\subsection{Existing works and Limitations}
\vspace{-0.1in}
At its core, model merging involves combining the parameters of pre-trained models to create a unified system that integrates and preserves the knowledge encoded in the original models. By operating directly on model parameters, model merging circumvents the need for retraining from scratch, offering a more efficient and secured alternative for the scenarios discussed above. 
\begin{table*}[ht]
\vspace{-0.05in}
\centering
\scalebox{0.8}{
\begin{tabular}{l|cc|cc}
\toprule
\textbf{Method} & \multicolumn{2}{c|}{\textbf{Inhibiting Properties}} & \multicolumn{2}{c}{\textbf{Undesirable Properties}} \\ 
\cmidrule{2-5}
& \textbf{Same Init. State} & \textbf{Training Labels} & \textbf{Model Inflation} & \textbf{Numerical Optimization} \\ 
\midrule
Weight Averaging & \textcolor{ForestGreen}{\ding{55}}  & \textcolor{ForestGreen}{\ding{55}} & \textcolor{ForestGreen}{\ding{55}} & \textcolor{ForestGreen}{\ding{55}} \\
Task Arithmetic~\citep{ilharco2023editingtaskarithmetic} & \textcolor{red}{\ding{51}}  & \textcolor{ForestGreen}{\ding{55}} & \textcolor{ForestGreen}{\ding{55}} & \textcolor{ForestGreen}{\ding{55}} \\
TIES~\citep{yadav2023tiesmerging} & \textcolor{red}{\ding{51}}  & \textcolor{ForestGreen}{\ding{55}} & \textcolor{ForestGreen}{\ding{55}} & \textcolor{ForestGreen}{\ding{55}} \\
\git~\citep{ainsworth2023gitrebasin} & \textcolor{ForestGreen}{\ding{55}}  & \textcolor{ForestGreen}{\ding{55}} & \textcolor{ForestGreen}{\ding{55}} & \textcolor{ForestGreen}{\ding{55}} \\
\permute~\citep{entezari2022the} & \textcolor{ForestGreen}{\ding{55}} & \textcolor{ForestGreen}{\ding{55}} & \textcolor{ForestGreen}{\ding{55}} & \textcolor{ForestGreen}{\ding{55}} \\
\zipit~\citep{stoica2024zipit} & \textcolor{ForestGreen}{\ding{55}} & \textcolor{ForestGreen}{\ding{55}} & \textcolor{red}{\ding{51}}\footnotemark[1] & \textcolor{ForestGreen}{\ding{55}} \\
AdaMerging~\citep{yang2024adamerging} & \textcolor{red}{\ding{51}}  & \textcolor{ForestGreen}{\ding{55}} & \textcolor{ForestGreen}{\ding{55}} & \textcolor{red}{\ding{51}} \\
RegMean~\citep{jin2023datalessregmean} & \textcolor{red}{\ding{51}}  & \textcolor{ForestGreen}{\ding{55}} & \textcolor{ForestGreen}{\ding{55}} & \textcolor{ForestGreen}{\ding{55}} \\
Fisher Merging~\citep{matena2022mergingfishermerging} & \textcolor{ForestGreen}{\ding{55}}  & \textcolor{red}{\ding{51}} & \textcolor{ForestGreen}{\ding{55}} & \textcolor{ForestGreen}{\ding{55}} \\
UQ-Merge~\citep{daheim2024modeluqmerging} & \textcolor{red}{\ding{51}}  & \textcolor{red}{\ding{51}} & \textcolor{ForestGreen}{\ding{55}} & \textcolor{ForestGreen}{\ding{55}} \\
EMR-Merging~\citep{emr-merging-DBLP:journals/corr/abs-2405-17461} & \textcolor{red}{\ding{51}}  & \textcolor{ForestGreen}{\ding{55}} & \textcolor{red}{\ding{51}} & \textcolor{ForestGreen}{\ding{55}} \\
\surgery~\citep{yang2024representationsurgery} & \textcolor{red}{\ding{51}}  & \textcolor{ForestGreen}{\ding{55}} & \textcolor{red}{\ding{51}} & \textcolor{red}{\ding{51}} \\
\name & \textcolor{ForestGreen}{\ding{55}}  & \textcolor{ForestGreen}{\ding{55}} & \textcolor{ForestGreen}{\ding{55}} & \textcolor{red}{\ding{51}} \\
\name{}++ & \textcolor{ForestGreen}{\ding{55}}  & \textcolor{ForestGreen}{\ding{55}} & \textcolor{ForestGreen}{\ding{55}} & \textcolor{ForestGreen}{\ding{55}} \\
\bottomrule
\end{tabular}}
\vspace{-0.1in}
\caption{\small Characterization of existing algorithms for model merging: \textcolor{red}{\ding{51}} denotes the presence of an undesirable property, whereas \textcolor{ForestGreen}{\ding{55}} indicates its absence. While a numerical learning-based optimization can be an effective model merging procedure, it also results in higher computational costs. In this context, a \textcolor{red}{\ding{51}} specifically highlights this increased computational burden.}

\label{tab:baselines}
\vspace{-0.2in}
\end{table*} In this work, we focus on model merging for graph neural networks (\gnn{}s). While several works on model merging exist, they are tailored for vision and language models. Consequently, when applied in the context of merging \gnns, unique challenges surface, which existing techniques fail to address adequately. Table~\ref{tab:baselines} summarizes these limitations, which we discuss below in detail. 

\begin{itemize}
\item \textbf{Assumption of shared initialization:} Most model-merging algorithms rely on the assumption that the models to be merged share a common initialization, often originating from a shared pre-trained foundation model. However, this assumption presents significant challenges in the context of \gnns, where such foundation models and shared initializations are rare, leading to difficulties in aligning model parameters effectively. 

\item \textbf{Shared dataset and tasks:} Many existing algorithms presume that the models being merged are trained on the same dataset and perform closely related tasks, such as classification over disjoint label sets. This assumption enables the merging process to exploit the alignment of models residing in different basins of the same task's loss landscape. However, when models are trained on diverse tasks, such as node classification and link prediction, with non-overlapping loss basins, the performance of these algorithms deteriorates, as they cannot reconcile the disparities in the underlying objective spaces.

\item \textbf{Model inflation:} The number of parameters in a model directly impacts its computational efficiency and GPU memory requirements. Ideally, the merged model should maintain the same size as the individual models being merged to preserve these efficiencies. However, several existing algorithms fail to meet this desideratum, leading to inflated model size with increased resource demands. Inflation may happen due to various design choices, such as the injection of adapter layers between model layers~\citet{yang2024representationsurgery} or partial merging of layers to avoid  degradation of performance~\citet{stoica2024zipit}.

\item \textbf{Numerical learning-based merging:} Merging algorithms can broadly be divided into two categories. The first category employs analytical operations on the input model parameters to produce the merged model. The second category adopts a numerical learning-based approach, optimizing the merged model's parameters by minimizing a loss function. While this method achieves better accuracy, it compromises on computational efficiency. 

\end{itemize}
\vspace{-0.1in}
\vspace{-0.1in}
\footnotetext[1]{\zipit employs ``partial zipping'', leaving some layers unmerged. The unmerged portion retains the original model layers, effectively doubling the parameter size for those layers. Empirical results demonstrate significant performance degradation when attempting to merge all layers.}
\subsection{Contributions}
\vspace{-0.05in}
 In this work, we present \name to address the above-outlined limitations. Our contributions are the following:

\begin{itemize}
    \item \textbf{Novel problem:} To the best of our knowledge, this is the first study surfacing the limitations of generic model merging algorithms for \gnns, underscoring the need for approaches specifically tailored to \gnns.
    \item \textbf{Task-agnostic algorithm design:} Regardless of the task, \gnns operate at the granularity of node embeddings, with task-specific aggregations performed post node embedding layers. Hence, if the merged model can preserve the node embeddings produced by the individual models being merged, it will remain effective on both tasks, even without having explicit knowledge of the tasks themselves. This core observation empowers our optimization objective for merging \gnns.
    \item {\bf Analytical solution:} We establish that our proposed optimization objective, when applied to message passing \gnns (\mpnn{}s), such as \gcn~\citep{kipf2017semi-gcn}, \gin~\citep{xu2018how}, \gat~\citep{velickovic2018graph} or \sage~\citep{graphsage}, allows reduction to an analytical solution. Consequently, the merged model can be obtained directly, negating the need for parameter optimization, enabling both efficiency and accuracy.
    \item {\bf Empirical benchmarking:} We present the first benchmarking study for model merging in \gnns and empirically establish that current state-of-the-art methods are ineffective on \gnns. In contrast, our embedding alignment objective with its analytical implementation delivers superior accuracy and achieves up to \textbf{136x} times speed-up compared to retraining from scratch.
\end{itemize}
\vspace{-0.1in}
\section{Problem Formulation}
\vspace{-0.1in}
\label{sec:formulation}
\begin{figure*}[t]
    \centering
    \vspace{-0.1in}
    \includegraphics[width=4.5in]{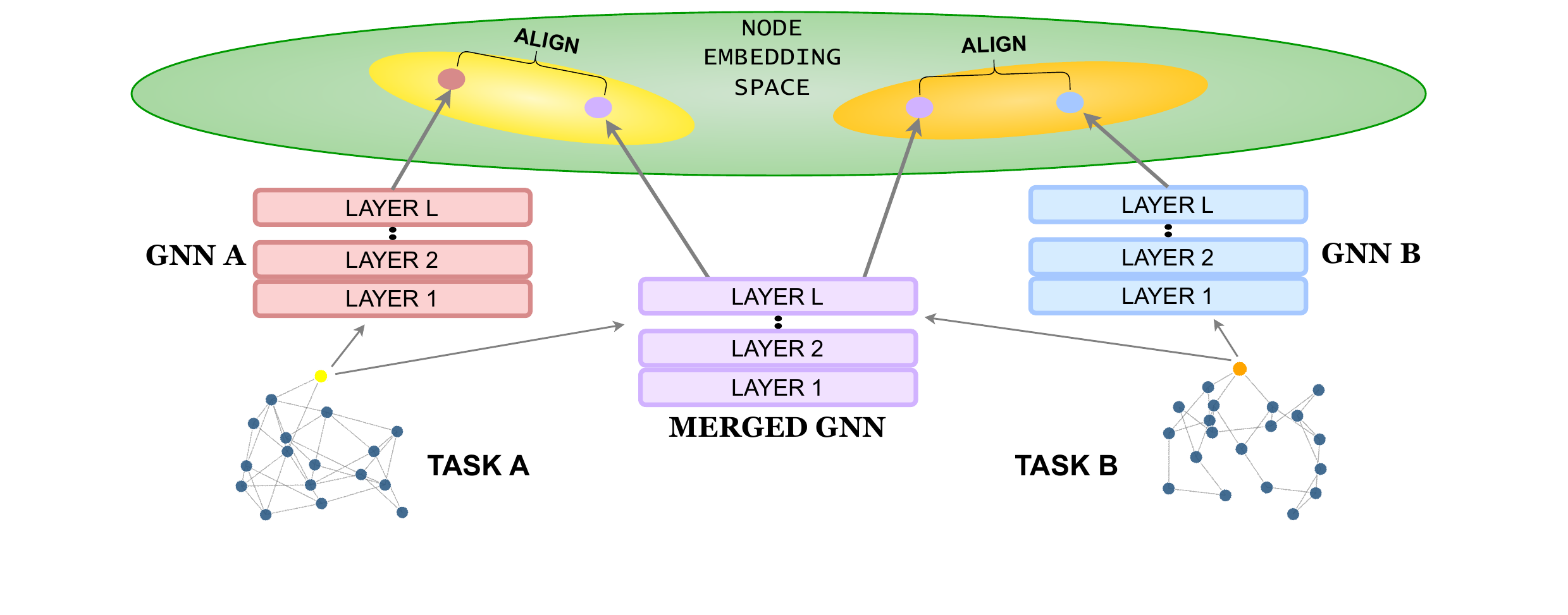} 
    \vspace{-0.4in}
    \caption{\small A visual depiction of the alignment objective in \name. The yellow and orange ellipses represent the regions where the highlighted nodes receive the correct prediction. \name aims to embed the nodes closer to their original embeddings, increasing the likelihood that the new embeddings fall within the ellipses. As stated in Prob.~\ref{prob:merge}, the merging graph(s) need not be the training graph or rely on supervision labels. While we assume a common graph for aligning base models, task-specific graphs can be used if needed.}
    \vspace{-0.1in}
    \label{fig:main_fig}
    \vspace{-0.1in}
\end{figure*}
\begin{defn}[Graph] \textit{Let $\CG = (\CV, \CE, \X)$ denote a graph over node set $\CV$ and edge set $\CE:\CV\times\CV$. $\X \in \mathbb{R}^{|V| \times |d|}$ denotes the node attributes encoded using $d$-dimensional feature vectors. The feature vector for a particular node $v\in\CV$ is denoted by $\cx_v$.}
\end{defn}
\vspace{-0.1in}
\textit{Prediction tasks} on graphs encompass diverse objectives, including node classification, link prediction, and node regression~\citep{graphsage}. We formally define the process of \textit{learning} a task using a \gnn as follows:
\begin{defn}[Learning a task]
\textit{For a prediction task $\CT$, let $\langle \mathbb{T},\CY\rangle$ be a training dataset where $|\mathbb{T}| = |\CY|$. Here, $\mathbb{T}$ contains task-relevant graph components and $\CY$ contains their corresponding ground-truth labels. A \gnn with parameters $\Theta$ is trained to minimize a loss function $\CL(\CY, \Theta(\mathbb{T}))$, optimizing the agreement between predictions and ground-truth labels such that $\CY \approx \Theta(\mathbb{T})$.}
\end{defn}
\vspace{-0.1in}
For node classification or regression, the components in $\mathbb{T}$ are nodes, while for link prediction, they correspond to edges. Similarly, the ground-truth labels in $\CY$ indicate class labels for node classification and the presence or absence of edges for link prediction. Commonly used loss functions include cross-entropy, negative log-likelihood, and RMSE.
The problem of model merging is now defined as follows.
\begin{prob}[Model Merging]\label{prob:merge}
\textit{Given $n$ \gnn models $\Theta_1, \Theta_2, \dots, \Theta_n$, the goal of model merging is to construct a merged model $\Theta_M$ such that, for any given graph $\CG$, the output of the merged model closely matches the outputs of the individual models. This can be formulated as minimizing the following objective:}
\vspace{-0.05in}
\begin{equation}
\small
\label{eq:mergeloss}
\frac{1}{n} \sum_{t=1}^n \CL_t(\Theta_t(\CG), \Theta_M(\CG)),
\end{equation}
\vspace{-0.2in}

\textit{where $\CL_t(\cdot, \cdot)$ represents the loss function corresponding to the model $\Theta_t$ for task $t$.}
\end{prob}
\vspace{-0.1in}
The loss function $\CL_t(\cdot, \cdot)$ quantifies the similarity between the predictions of $\Theta_t$ and $\Theta_M$. In scenarios where the tasks differ, the scales of the loss functions across models might vary, necessitating normalization to ensure comparability. However, for simplicity and clarity of exposition, we omit such normalization factors in this formulation.

In addition to the objective in Prob.~\ref{prob:merge}, the following desiderata are crucial:
\vspace{-0.05in}
\begin{itemize}
\item \textbf{Computational Efficiency:} The merging process should be significantly faster than training the merged model $\Theta_M$ from scratch using the combined training data of all individual models.
\item \textbf{Independence from Labeled Data:} The merging process should rely solely on the parameters of the individual models and any inference data, without requiring access to the original training data or its ground-truth labels.
\item \textbf{Model size:} The number of parameters in the merged model $\Theta_M$ should be the same as that of any of the individual models $\Theta_i$
\end{itemize}
\vspace{-0.05in}
We assume that the models being merged belong to the same \gnn architecture. This assumption aligns with existing model merging algorithms, as merging models with heterogeneous architectures remains an open problem. 
\vspace{-0.1in}
\section{\name: Proposed Methodology}
\vspace{-0.1in}
\name leverages the insight that \gnn layers universally compute node embeddings, regardless of the specific task. Therefore, if the merged model can replicate the node embeddings generated by each base model, it can also replicate their outputs. To achieve this, we first define an optimization objective focused on preserving the node embeddings from the base models within the merged model. This objective is then relaxed to facilitate an analytical solution and enable various computational optimizations. The following subsections outline these steps in detail.
\vspace{-0.1in}
\subsection{Computation Framework of \gnns}
\label{sec:gnn}
\vspace{-0.1in}
\gnns update node embeddings of the input graph in a layer-by-layer manner. The $0^{th}$ layer embedding of node $v\in \CV$ is simply $\mathbf{h_v^0 = x_v}$. In layer $\ell$, each node $v$ draws messages from its neighbors $\CN_v=\{u\in\CV\mid(u,v)\in\CE\}$\footnote{In a graph transformer, messages are drawn from all nodes in a graph. The proposed framework trivially extends to this setting since it is simply an extension of the neighborhood definition.}. The message drawn by node $v$ from its neighbor $u$ is simply the embedding of $u$ in layer $\ell-1$, denoted as $\ch_u^{\ell-1}$. The messages are then aggregated using either some predefined function (e.g., \textsc{MeanPool}) or neural networks (e.g., \gat~\citep{velickovic2018graph}). 
\vspace{-0.05in}
\begin{align}
\small
\label{eq:aggregate}
\cm_v^\ell &= \text{\textsc{Aggregate}}^\ell(\CS_v^{\ell}) \\
\text{where, }\CS_{v}^{\ell}&=\{\!\!\{\ch_u^{\ell-1}, \forall u \in \mathcal{N}_v\}\!\!\}
\end{align}
\vspace{-0.3in}

Here, $\CS_{v}^{\ell}$ represents the multiset of messages drawn from the neighbors. 
  The $\ell^{th}$ layer embedding of node $v\in\CV$ is then obtained by combining the aggregated message with $v$'s own embedding and then passing it through an \mlp. Formally, this may be denoted as:
  \vspace{-0.05in}
\begin{equation}
    \small
    \mathbf{h}_v^{\ell} = \mlp\left(\textsc{Combine}^{\ell}\left(\mathbf{h_v^{\ell-1}},\cm_v^\ell\right)\right)
\end{equation}
Here, $\textsc{Combine}^{\ell}$ is another pre-defined function. As examples, while \sage concatenates learnable linear transformations on $\mathbf{h_v^{\ell-1}}$ and $\cm_v^\ell$, \gcn and \gat add self-loops to $v$ and then use degree-weighting and learnable attention-weighted \textsc{SumPool} respectively.

\vspace{-0.1in}
\subsection{Merging through Node Embedding Alignment}
\vspace{-0.1in}
The prediction from a \gnn is a function of the node embeddings. Hence, even if the model parameters of the merged model are distinctly dissimilar to the base models, as long as the embeddings produced are similar, the outputs would be similar. Grounded on this observation, we shift the focus from combining models in the parameter space to optimizing them with respect to the embedding space. Fig.~\ref{fig:main_fig} visually illustrates the idea. 
 Formally, we propose a node embedding alignment objective as follows.

Let $\Theta_1,\cdots,\Theta_n$ 
 be the base models being merged. Let $\CG(\CV,\CE,\X)$ be a graph from the same domain where the merged model will be applied. Note that  $\CG$ need not be the train graph. 
  Under the node alignment objective, for each \gnn layer, we aim to align the embeddings produced by the merged model $\Theta_M$ on $\CG$ with the embeddings  produced by each of the base models $\Theta_i,\;1\leq i \leq n$ on $\CG$. 
  More concretely, we minimize:
\vspace{-0.1in}
\begin{alignat}{2}
\small
\label{eq:opt}
\overbrace{\sum_{i=1}^n}^{\substack{\text{For each} \\ \text{base model,}}} \overbrace{\sum_{\ell=1}^L}^{\substack{\text{\gnn}\\ \text{layer}}}\overbrace{\sum_{\forall v \in \CV}}^{\text{node}}&\|\Theta_M^{\ell}\left(\ch_{v,M}^{\ell-1},\CS_{v,M}^{\ell-1}\right) - \Theta_i^{\ell}\left(\ch_{v,i}^{\ell-1},\CS_{v,i}^{\ell-1}\right)\|_2
\end{alignat}
\vspace{-0.1in}

Here, $\ch_{v,i}^{\ell-1}$ represents the embedding of node $v$ and $\CS_{v,i}^{\ell-1}$ denotes the embeddings of its neighbors in model $\Theta_i$ from layer $\ell-1$ (analogously defined for $\Theta_M$). The parameters $\Theta_{i}^{\ell}$ are responsible for the transformations at layer $\ell$. Our objective is to determine the parameters of $\Theta_{M}^{\ell}$ for each layer $\ell$, ensuring that the merged model $\Theta_M$ generates embeddings that closely align with those produced by the base models.
Although the minimization task described in Eq.~\ref{eq:opt} is both task-agnostic and independent of training labels, it is computationally intensive. The process is equivalent to training a \textit{student} \gnn ($\Theta_M$) to mimic a set of \textit{teacher} \gnns (the base models). This approach contradicts the computational efficiency requirements outlined in \S~\ref{sec:formulation}. However, we will demonstrate how a minor relaxation of the formulation can dramatically reduce the computational burden, aligning with our efficiency goals.

\vspace{-0.1in}
\subsection{Independent Node Embedding Alignment}
\label{sec:ind}
\vspace{-0.1in}
Learning the parameters minimizing Eq.\ref{eq:opt} is expensive since the merging process at layer $\ell$ depends on all preceding layers since its input is determined by the outputs of layers $1$ to $\ell-1$. This dependency necessitates backpropagation of gradients through multiple layers, increasing computational overhead and slowing convergence. 

To ease the computational burden without any significant disruption on our objective, we introduce a slight relaxation. Instead of aligning the node embeddings produced by the merged model as a whole, we align the layer-wise node embeddings independently of each other. 
Specifically, instead of sending $\ch_{v,M}^{\ell-1}$ and $\cS_{v,M}^{\ell-1}$ to $\Theta_M^{\ell}$, we directly send $\ch_{v,i}^{\ell-1}$ and $\cS_{v,i}^{\ell-1}$. Consequently, the minimization objective reduces to:
\vspace{-0.1in}
\begin{equation}
\small
\label{eq:optrelax}
\sum_{i=1}^n\sum_{\ell=1}^L\sum_{\forall v \in \CV}\|\Theta_M^{\ell}\left(\ch_{v,i}^{\ell-1},\CS_{v,i}^{\ell-1}\right) - \Theta_i^{\ell}\left(\ch_{v,i}^{\ell-1},\CS_{v,i}^{\ell-1}\right)\|_2 
\end{equation}
The key insight behind this relaxation is that the learning objective for $\Theta_M^{\ell}$ becomes independent of the preceding layers of the merged model since its input is no longer derived from the outputs of its own previous layers. Intuitively, this adjustment directs each layer of the merged model toward a parameter space where the linear transformations applied to the node embeddings of the base models (rather than its own embeddings) closely approximate the transformations induced by the base model's parameters. 
This relaxation is expected to have a mild effect since, at layer $0$, all models share the same input, i.e., $\ch^0_v = \cx_v$. Consequently, for a 1-layer \gnn, the relaxed objective in Eq.~\ref{eq:optrelax} is equivalent to the original objective in Eq.~\ref{eq:opt}. For deeper \gnns, if $\Theta_M^1$ effectively approximates the base models, then $\ch_{v,M}^{\ell} \approx \ch_{v,i}^{\ell}$ and $\cS_{v,M}^{\ell} \approx \cS_{v,i}^{\ell}$, resulting in transitive consistency across subsequent layers. We further note that \gnns are typically not deep (often $\leq 3$ layers) due to the well-established problems of oversquashing and oversmoothing~\citep{oversmoothing, oversquashing}. Next, we demonstrate how this relaxation enables analytical solutions for popular \gnn architectures, resulting in dramatic efficiency improvements.

\vspace{-0.1in}
\subsection{Analytical Solution}
\vspace{-0.1in}
\label{sec:cf}
In any layer of a \gnn, the operations can be categorized into two types: (1) non-learnable aggregations (e.g., \textsc{SumPool}) and (2) learnable transformations (such as an \mlp or attention computation). 


The learnable parameters are solely associated with such linear transformations. Let us denote the learnable weight matrices in layer $\ell$ for model $\Theta_{i}$ as $\cW_{1,i}^{\ell},\cdots,\cW_{K,i}^{\ell}$, where $K$ is the total number of transformations conducted in any layer $\ell$. Similarly, the vectors on which these transformations are applied for model $\Theta_{i}$ are denoted as $\cz_{v,1,i}^{\ell-1},\cdots,\cz_{v,K,i}^{\ell-1}$. The outputs of these transformations are denoted as $\cg_{v,1,i}^{\ell},\cdots,\cg_{v,K,i}^{\ell}$. Note that in a \gnn, while the weight matrices are shared across all nodes, the embeddings on which they operate are node-specific.

Since all parameters are associated with linear transformations only in a \gnn, Eq.~\ref{eq:optrelax} can be re-written as:
\vspace{-0.1in}
\begin{equation}
\small
\label{eq:cf1}
\sum_{i=1}^n\sum_{\ell=1}^L\sum_{k=1}^K\sum_{\forall v \in \CV}\|\cz_{(v,k,i)}^{\ell-1}\cW_{k,M}^{\ell} - \cg_{v,K,i}^{\ell}\|_2 
\end{equation}
 Since each linear transform for each layer happens independently, minimizing Eq.~\ref{eq:cf1} is equivalent to optimising each $\cW_{k,M}^{\ell}$ as follows:
 \vspace{-0.1in}
\begin{equation}    
\label{eq:cf2}
\min\limits_{\cW_{k,M}^{\ell}} \sum_{i=1}^n\sum_{\forall v \in \CV}\|\cz_{(v,k,i)}^{\ell-1}\cW_{k,M}^{\ell} - \cg_{v,K,i}^{\ell}\|_2 
\end{equation}
Let $\CZ_{k,i}^{\ell-1}$ be the matrix containing $\cz_{(v,k,i)}^{\ell-1}$ and $\cG_{k,i}^{\ell}$ be the matrix containing $\cg_{(v,k,i)}^{\ell}$  $\forall v \in \CV$. Eq.~\ref{eq:cf2} can then be re-written as:
\vspace{-0.1in}
\begin{equation}
\small
\label{eq:cf3}
    \min\limits_{\cW_{k,M}^{\ell}} \sum_{i=1}^n \|\CZ_{k,i}^{\ell-1}\cW_{k,M}^{\ell} - \cG_{K,i}^{\ell}\|_F^2
\end{equation}
were $\|A\|^2_F$ represents the frobenius norm of matrix $A$.
Since Eq.~\ref{eq:cf3} is convex, the minima is achieved when the gradient with respect to $\cW_{k,M}^{\ell}$ is $0$. Setting it to 0 and solving for $\cW_{k,M}^{\ell}$, we get:
\vspace{-0.05in}
\begin{equation}
\small
(\cW_{k,M}^{\ell})^{\mathtt{T}}  = \sum_{i=1}^n(\cG_{K,i}^{\ell})^{\mathtt{T}}\cG_{K,i}^{\ell}(\sum_{i=1}^n(\CZ_{k,i}^{\ell-1})^{\mathtt{T}}\CZ_{k,i}^{\ell-1})^{-1}
\label{eq:cfw}
\end{equation}
Eq.~\ref{eq:cfw} presents an analytical solution to directly compute the weights of the merged \mpnn layers, which minimises the desired objective function (Eq.~\ref{eq:optrelax}).

\textbf{Efficiency implications:} Since each layer is merged independently, the complexity of the training process reduces, allowing for simpler weight adjustments and gradient computations. Furthermore, owing to independence, each layer can be merged in an embarrassingly parallel fashion.   
\vspace{-0.1in}
\subsubsection{Illustrative example: Applying  analytical framework to \gcn}
\label{sec:gcn}
\vspace{-0.05in}
As an illustrative example, we apply the above result in the context of \gcn. In App.~\ref{app:derivation}, we present analytical versions for other popular \mpnn architectures, including  \gin, \gat and \sage.

The node embedding update equation for \gcn is: 
\vspace{-0.05in}
\begin{equation}
\small
\label{eq:gcn}
  \mathbf{h}_v^{(\ell)} = \sigma\left(\sum_{u \in \mathcal{N}_v \cup \{i\}} \frac{1}{\sqrt{d_u d_v}} \mathbf{h}_u^{(\ell-1)} \mathbf{W}^{(\ell)}\right)  
\end{equation}
where $d_v$ denotes the degree of node $v$ (including a self-loop), $\sigma$ is an activation function, such as \relu{} and $\cW^{\ell}$ is a learnable weight matrix. Hence, when applied to the generic framework expressed in Eq.~\ref{eq:cfw}, $K=1$, i.e., there is only one learnable weight matrix per layer. Now, to compute $\cW^{\ell}_{1,M}$ using Eq.~\ref{eq:cfw}, we need to know $\cG^{\ell}_{1,i}=\{\cg_{v,1,i}^{\ell}\mid v\in\CV\}$ and $\CZ^{\ell-1}_{1,i}=\{\cz_{v,1,i}^{\ell-1}\mid v\in\CV\}$. From Eq.~\ref{eq:gcn}, it is easy to see that for any \gcn model $\Theta_i$, we have:
\vspace{-0.05in}
\begin{equation}
\small
 \mathbf{g}_{v,1,i}^{\ell} = \underbrace{\left(\sum_{j \in \mathcal{N}_v \cup \{i\}} \frac{1}{\sqrt{d_v d_u}} \mathbf{h}_{u,1,i}^{(\ell-1)} \right)}_{\cz_{v,1,i}^{\ell-1}}\mathbf{W}^{(\ell)}_{1,i}    
\end{equation}
\begin{table*}[t]
\centering
\renewcommand{\arraystretch}{1} 
\scalebox{0.8}{
\begin{tabular}{l|cccccccccc}
\toprule
 \textbf{Dataset} & \textbf{M} & \textbf{Raw} & \textbf{\wa} & \textbf{\git} & \textbf{\permute} & \textbf{\zipit} & \textbf{\surgery} & \textbf{\name} & \textbf{\name{}++} \\ \midrule
\multirow{2}{*}{\textbf{Arxiv}} & Model 1 & 80.85 & 68.56 & 65.83 & 70.56 & 69.53 & 68.41 & 75.39 &\cellcolor{Gray}77.47 \\ 
                                & Model 2 & 82.51 & 42.09 & 10.48 & 58.34 & 57.21 & 66.41 & \cellcolor{Gray}74.02 &73.02 \\ 
                                \midrule
\multirow{2}{*}{\textbf{AmzComp}} & Model 1 & 95.81 & 64.34 & 65.96 & 71.03 & 73.07 & 85.62 & 92.58 &\cellcolor{Gray}93.94 \\ 
                                & Model 2 & 93.09 & 79.42 & 49.08 & 85.42 & 81.35 & 78.45 & \cellcolor{Gray}92.83 &92.18 \\ 
                                \midrule
\multirow{2}{*}{\textbf{AmzPhoto}} & Model 1 & 94.72 & 63.10 & 70.09 & 70.22 & 88.21 & 85.37 & 93.93 &\cellcolor{Gray}94.72 \\ 
                                & Model 2 & 94.81 & 68.61 & 39.16 & 81.12 & 73.53 & 90.89 & 94.01 &\cellcolor{Gray}94.81 \\ 
                                \midrule
\multirow{2}{*}{\textbf{Cora}} & Model 1 & 86.54 & 67.10 & 84.50 & 75.75 & 67.02 & \cellcolor{Gray}87.13 & 84.32 &85.38 \\ 
                                & Model 2 & 93.03 & 92.72 & 48.10 & 89.71 & 88.33 & 89.43 & 91.08 &\cellcolor{Gray}93.35 \\ 
                                \midrule
\multirow{2}{*}{\textbf{Reddit}} & Model 1 & 97.59 & 88.38 & 92.10 & 92.33 & 90.01 & 83.59 & \cellcolor{Gray}96.82 &96.80 \\ 
                                & Model 2 & 94.35 & 48.23 & 05.82 & 72.21 & 74.03 & 74.58 & \cellcolor{Gray}94.33 & 94.30 \\ 
                                \midrule
\multirow{2}{*}{\textbf{WikiCS}} & Model 1 & 86.63 & 59.31 & 81.09 & 62.41 & 69.92 & 68.07 & 86.01 & \cellcolor{Gray}86.79 \\ 
                                & Model 2 & 85.95 & 42.64 & 28.54 & 62.21 & 69.71 & 58.25 & 82.50 &\cellcolor{Gray}84.54 \\ 
                                \midrule
\multicolumn{2}{c}{\textbf{Average}} & 90.49 & 65.37 & 53.39 & 74.27 & 75.16 & 78.01 & \textbf{88.14} & \cellcolor{Gray}\textbf{88.94} \\ \bottomrule
\end{tabular}}
\vspace{-5pt}
\caption{\small \textbf{In-domain Dataset Results.} \name and \name{}++ compared with baselines on merging models trained on disjoint label splits of the same dataset. The best results on each dataset-model pair are shaded. Metric reported: Accuracy $(\%)$.}
\label{tab:results1}
\vspace{-0.2in}
\end{table*}
\vspace{-0.3in}
\section{Experiments}
\vspace{-0.15in}
In this section, we benchmark \name and establish:
\vspace{-0.05in}
\begin{itemize}
    \item \textbf{Efficacy in the context of \gnns:} This work presents the first benchmarking study of model-merging algorithms for \gnns, revealing significant performance deterioration in merged models. These findings highlight the necessity of a specialized algorithm tailored for \gnns. \name addresses this critical gap, outperforming state-of-the-art model-merging algorithms in the context of \gnns.
    \item \textbf{Efficiency:} \name is \textbf{136x} times faster than training a model from scratch, with only minor drops in performance. This efficiency is achieved by leveraging an analytical solution to compute the weights of the merged model, which we derive by carefully analyzing the message-passing aggregation of \gnns.
\end{itemize}
\vspace{-0.05in}
The implementation of our algorithm is available at \url{https://anonymous.4open.science/r/Model-Merging-GNNs-4C55}.
\vspace{-0.1in}
\subsection{Experimental Setup}
\vspace{-0.05in}
\label{sec:setup}
\vspace{-0.05in}
The details of our hardware and software environment are listed in App.~\ref{app:experiments}.

\textbf{Tasks:} We benchmark \name on three types of model merging scenarios:
\vspace{-0.05in}
\begin{enumerate}
    \item \textbf{Node Classification on In-domain Datasets:}\label{sec:in-domain}   
    We create two disjoint label splits from the same dataset and train a model on each split independently. The merging process is then performed on these two models, simulating a scenario where new labels are introduced after the initial training.
    \item \textbf{Node Classification on Different Datasets:}\label{sec:diff_datasets}  
    Given $N$ models, each trained for node classification on a distinct dataset, we merge these models into a single unified model. The performance of the merged model is subsequently evaluated on the test sets of the respective datasets.
    To ensure a common architecture for \gnn models across different datasets, we only the utilize Text-Attributed Graphs from Table~\ref{tab:datasets1} (first five rows). Raw text attributes associated with nodes in these datasets are processed using Sentence-BERT~\citep{reimers2019sentence} to generate uniform feature representations.
    \item \textbf{Node Classification and Link Prediction on Different Datasets:}\label{sec:diff_tasks}
    We merge models trained on different tasks on different datasets.
\end{enumerate}

\textbf{Datasets.} Table~\ref{tab:datasets1} lists the $8$ graph datasets used for our experiments. \\
\textbf{Baselines.} All the existing algorithms listed in Table~\ref{tab:baselines} with a \textcolor{red}{\ding{51}} in the ``\textit{Inhibiting properties}'' column are inapplicable in our setting. These include methods requiring labeled data (e.g., Fisher Merging, UQ-Merge) or those dependent on a common pre-trained backbone fine-tuned for all tasks (e.g., Task Arithmetic, AdaMerging, etc.). After excluding such algorithms, we focus on the remaining applicable methods: Weight Averaging (\wa), \git, \permute, and \zipit. Additionally, we compare against \surgery, a post-hoc refinement method applied to a merged model. \surgery supports a variant where \wa is used to create the merged model, making it compatible with our setting. 
We use \name to denote the layer-independent learning-based methodology proposed in this work and \textbf{\name{}++} to denote the analytical version.\\
\textbf{Architectures.} While our main results are presented on \gcn, we also evaluate generalizations to \sage and \node~\citep{wu2022nodeformer}, a graph transformer.
\vspace{-0.1in}
\subsection{Results}
\vspace{-0.05in}
\label{sec:main_res}
\vspace{-0.05in}
\textbf{In-domain Datasets.} In Table~\ref{tab:results1}, we present the results of merging \gcn{}s trained on disjoint label splits (of equal sizes) for node classification tasks on the same dataset. Both \name and \name{}++ demonstrate an average accuracy comparable to that of the base models, showcasing the effectiveness of our method. Notably, \name{}++ achieves a significant improvement over existing methods, outperforming \surgery by $\mathbf{10.93\%}$, \zipit by $\mathbf{13.78\%}$, \permute by $\mathbf{14.67\%}$, \git by $\mathbf{35.53\%}$, and weight averaging by $\mathbf{23.57\%}$.
\begin{table*}[t]
\centering
\scalebox{0.75}{
\begin{tabular}{l|cccccccc}
\toprule
\textbf{Datasets} & \textbf{Raw}& \textbf{\wa} & \textbf{\git} & \textbf{\permute} & \textbf{\zipit} & \textbf{\surgery} & \textbf{\name} & \textbf{\name{}++} \\
\midrule
\textbf{Citeseer} & 81.97   & 78.09    & 80.25    & 79.15    & 78.68    & 79.50    & \cellcolor{Gray}82.91    & 82.44    \\ 
\textbf{Pubmed}  & 79.02    & 75.94    & 22.23    & 78.47    & 77.25    & 68.69    & \cellcolor{Gray}79.14    & 79.04    \\ \midrule
\textbf{Citeseer} & 81.97   & 67.54    & 71.78    & 73.19    & 74.92    & 79.56    & 82.44    & \cellcolor{Gray}82.60    \\ 
\textbf{WikiCS}  & 79.32    & 60.27    & 22.90    & 61.99    & 63.28    & 71.19    & 78.00    & \cellcolor{Gray}78.21    \\ \midrule
\textbf{Arxiv}   & 73.10    & 68.43    & 53.12    & 53.56    & 50.11    & 60.46    & \cellcolor{Gray}72.21    & 71.98    \\ 
\textbf{WikiCS}  & 79.32    & 66.89    & 25.98    & 61.55    & 67.16    & 72.40    & \cellcolor{Gray}79.01    & 78.67    \\ \midrule
\textbf{Arxiv}   & 73.10    & 61.4     & 60.47    & 57.64    & 59.05    & 57.66    & 72.62    & \cellcolor{Gray}72.65    \\ 
\textbf{Pubmed}  & 79.02    & 74.28    & 20.88    & 78.04    & 78.12    & 75.39    & 79.08    & \cellcolor{Gray}79.13    \\ \midrule
\textbf{Pubmed}  & 79.02    & 76.20    & 67.88    & 75.81    & 75.16    & 74.97    & \cellcolor{Gray}78.96    & \cellcolor{Gray}78.96    \\ 
\textbf{WikiCS}  & 79.32    & 70.68    & 8.02     & 69.95    & 73.16    & 69.36    & \cellcolor{Gray}79.39    & 78.89    \\ \midrule
\textbf{Average} & 78.51 & 69.97 & 43.35 & 68.935 & 69.68 & 70.91 & \cellcolor{Gray}\textbf{78.37} & \textbf{78.25}\\ \bottomrule
\end{tabular}}
\vspace{-0.1in}
\caption{\small Merging of models trained on different datasets. Results on a larger number of dataset pairs are reported in Table~\ref{tab:results2} in the appendix. Metric reported: Accuracy $(\%)$}
\label{tab:results2half}
\vspace{-0.1in}
\end{table*}
\\\textbf{Different Datasets.} 
Table~\ref{tab:results2} presents the performance of merging \gcn{}s trained for node classification across two distinct datasets. Both \name and \name{}++ achieve average accuracies that are comparable to the base models. Additionally, \name outperforms the closest baseline by \textbf{7.02\%}. While \surgery is competitive in some cases, it must be noted that it comes at the cost of model inflation by introducing task-specific parameters. To further stress-test the methods, we extend the analysis by merging more than two models trained on multiple datasets. The full results of this evaluation are provided in Tables~\ref{tab:results3},~\ref{tab:results4}, and~\ref{tab:results5} in the appendix. 
In Fig.~\ref{fig:datasets}, we present the average accuracy of the merged models across all dataset combinations of a particular size (i.e., the row corresponding to ``Average'' in Tables~\ref{tab:results2}, ~\ref{tab:results3},~\ref{tab:results4}, and~\ref{tab:results5}). In this analysis, we have excluded \git since the source code does not support merging of more than two models, 
 and its performance is not competitive even for two datasets (Table~\ref{tab:results2}). Additionally, \name{}++ is omitted from Fig.~\ref{fig:datasets} since its performance closely mirrors that of \name (see Tables~\ref{tab:results2},~\ref{tab:results3},~\ref{tab:results4}, and~\ref{tab:results5} in the Appendix).
As depicted in Fig.~\ref{fig:datasets}, \name demonstrates significantly superior robustness when merging multiple models, exhibiting only a negligible decline in performance even when merging up to five models. Notably, when merging five models, \name achieves an impressive \textbf{23.53\%} improvement over the best baseline.\\
Results on generalization to \textbf{different architectures} and merging of \textbf{different tasks} are present in Appendix Section~\ref{sec:additional}.
\vspace{-0.1in}
\subsection{Ablation Study}
\label{sec:ablation}
\vspace{-0.1in}
\begin{wrapfigure}{r}{0.4\textwidth}
 \vspace{-0.5in}
    \centering
    \includegraphics[scale=0.6]{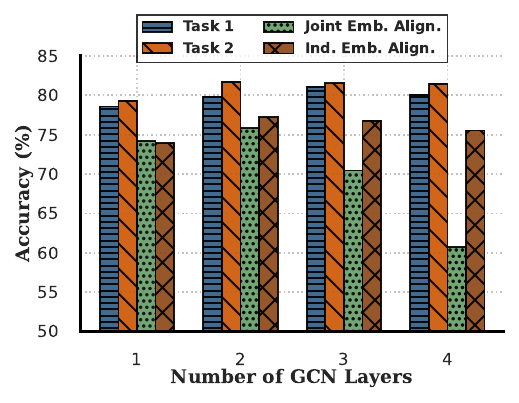}
    \vspace{-10pt} 
    \caption{\small Variation of performance of the two objective functions as the number of \gcn layers is changed for the arxiv dataset.}
    \label{fig:layers}
    \vspace{-0.15in}
\end{wrapfigure}
We aim to address two key questions in the next experiment. First, how does the number of \gnn layers impact model merging performance? Second, as discussed in \S~\ref{sec:ind}, learning parameters through \textit{joint} node embedding alignment introduces computational overhead and slower convergence (Eq.~\ref{eq:opt}). To mitigate this, we propose layer-wise independent node embedding alignment, which serves as a relaxation of the original objective (Eq.~\ref{eq:optrelax}). What effect does this relaxation have on performance? Fig.~\ref{fig:layers} presents the performance of the two optimization strategies as we vary the number of \gcn layers in the merging models on the arXiv dataset. A clear trend emerges: as the number of \gnn layers increases, the performance of the joint node alignment strategy deteriorates. In contrast, the relaxed optimization strategy, which aligns each layer independently, remains stable and does not suffer from this degradation. This behavior is attributed to the vanishing gradient problem becoming more pronounced in joint node alignment as the number of layers increases. Treating layers independently circumvents this issue, as the optimization problem remains decoupled from the depth of the network.
\begin{figure*}[t!]
    \centering
    \vspace{-0.2in}
    \subfloat[]{
    \label{fig:emb1}
    \includegraphics[width=0.25\textwidth]{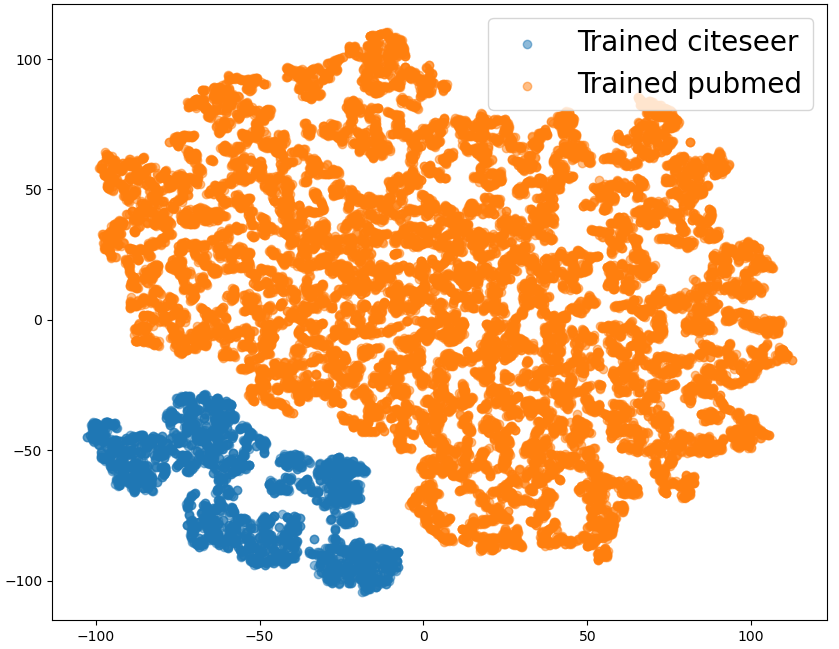}}
    \subfloat[]{
    \label{fig:emb2}
    \includegraphics[width=0.25\textwidth]{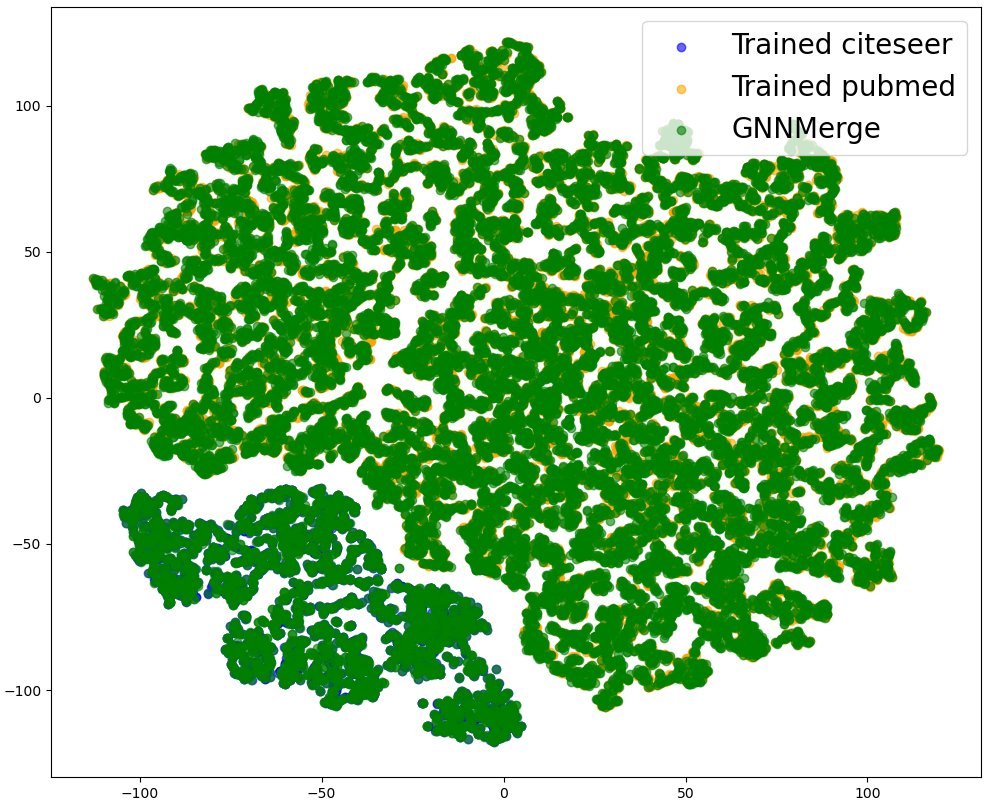}}
    \subfloat[]{
    \label{fig:emb3}
    \includegraphics[width=0.25\textwidth]{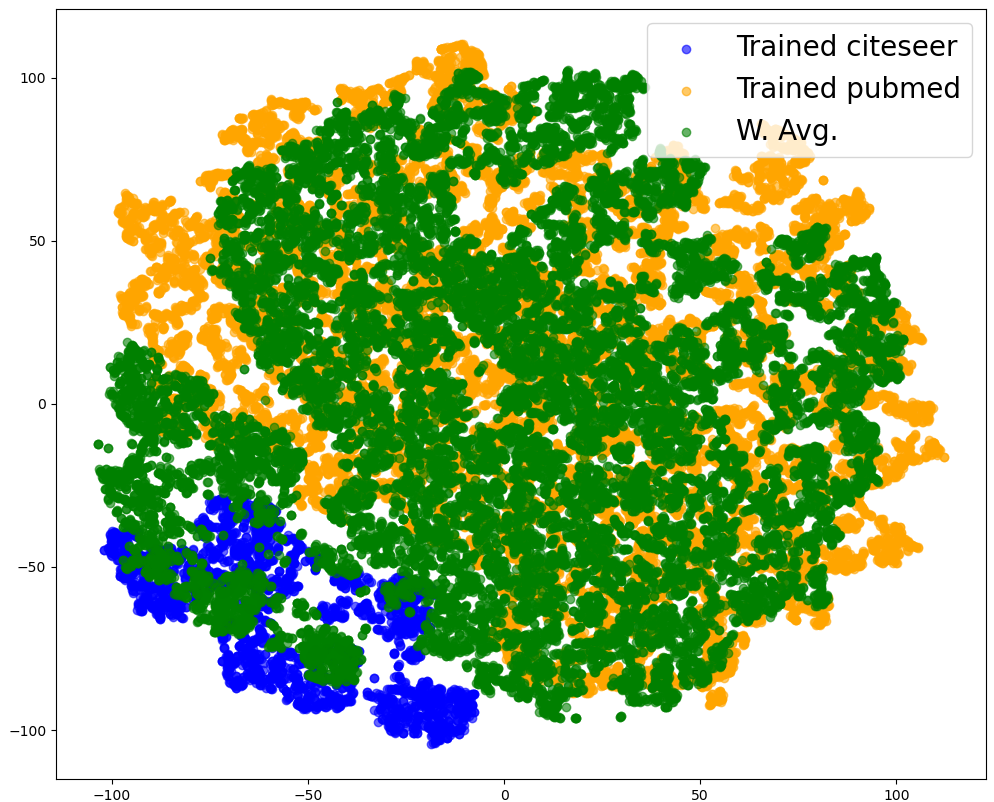}}
    \vspace{-6pt}
    \caption{\small Visual Illustration of embedding alignment using \name{} and \wa}
    \label{fig:three_figs}
    \vspace{-0.3in}
    
\end{figure*}
\vspace{-0.1in}
\subsection{Visual Analysis}
\label{sec:visual}
\vspace{-0.1in}
In this section, we investigate the effectiveness of our objective function in aligning the node embeddings as intended. Fig.~\ref{fig:emb1} presents the node embeddings for Citeseer and Pubmed, generated by their respective trained models, when projected to two dimensions using TSNE. Fig.~\ref{fig:emb2} and Fig.~\ref{fig:emb3} overlay the node embeddings produced by \name and \wa onto the embeddings from Fig.~\ref{fig:emb1} respectively. We observe that the embeddings generated by \name fully overlap with the base model's sembeddings. In contrast, in Fig.~\ref{fig:emb3}, there are patches without overlap with \wa embeddings. This visualization demonstrates \name{}'s superior ability to align embeddings with those produced by the base models, resulting in significantly improved performance. More analysis is present in App.~\ref{app:visual}.
\vspace{-0.15in}
\section{Conclusions}
\vspace{-0.15in}
In this work, we present the first comprehensive benchmarking of model merging algorithms for \gnns. Our analysis reveals that state-of-the-art merging techniques suffer significant performance degradation when applied to \gnns. To bridge this gap, we introduce \name, which employs a task-agnostic node embedding alignment strategy—preserving embeddings rather than directly merging model parameters. A key innovation is our analytical solution for message-passing \gnns, enabling direct merging without costly parameter optimization. Empirical results demonstrate that \name and its analytical variant, \name{}++, achieve up to $24\%$ higher accuracy than existing methods while delivering over two orders of magnitude speed-up compared to training from scratch. As the first work in this space, our approach paves the way for efficient and scalable model merging in GNNs, with potential applications in continual learning, multi-task graph-based AI systems, and privacy-preserving graph learning.
\newpage
\bibliography{./main}
\bibliographystyle{iclr2025_conference}

\newpage
\onecolumn
\appendix
\section{Analytical Derivation For Other GNNs}
\label{app:derivation}
\subsection{GraphSAGE}
The node embedding update equation for \textbf{GraphSAGE} is as follows:
\begin{equation}
\mathbf{h}_v^{(\ell)} = \sigma \left(\mathbf{h}_v^{(\ell-1)} \mathbf{W_1}^{(\ell)} \left| \right| \sum_{u \in \mathcal{N}_v} \frac{1}{\lvert \mathcal{N}_v \rvert} \mathbf{h}_u^{(\ell-1)} \mathbf{W_2}^{(\ell)}\right)
\end{equation}
where:
\begin{itemize}
    \item \(\mathcal{N}_v\): Set of neighbors of node \(i\).
    \item \(\mathbf{h}_u^{(\ell)}\): Feature vector of node \(u\) at layer \(l\).
    \item $\sigma$: Activation function like ReLU. 
    \item \(\mathbf{W_1}^{(\ell)},\mathbf{W_2}^{(\ell)}\): Trainable weight matrices at layer \(l\).
\end{itemize}

The trainable weight matrix \(\mathbf{W_2}^{(\ell)}\) can be factored out to obtain:
\begin{equation}
\mathbf{h}_v^{(\ell)} = \sigma \left(\mathbf{h}_v^{(\ell-1)} \mathbf{W_1}^{(\ell)} \left| \right| \left(\sum_{u \in \mathcal{N}_v} \frac{1}{\lvert \mathcal{N}_v \rvert} \mathbf{h}_u^{(\ell-1)}\right) \mathbf{W_2}^{(\ell)}\right)
\end{equation}

For a given target node, the term $\left(\sum_{u \in \mathcal{N}_v} \frac{1}{\lvert \mathcal{N}_v \rvert} \mathbf{h}_u^{(\ell-1)}\right)$ can be computed independent of $\mathbf{W_2}^{(\ell)}$ and can be denoted by $\mathbf{k}_v^{(\ell)}$. Hence, the node update equation becomes:
\begin{equation}
\label{eq:sage}
\mathbf{h}_v^{(\ell)} = \sigma \left(\mathbf{h}_v^{(\ell-1)} \mathbf{W_1}^{(\ell)}\left| \right| \mathbf{k}_v^{(\ell-1)}\mathbf{W_2}^{(\ell)}\right)
\end{equation}
Hence, when applied to the generic framework, $K=2$, i.e. there is only two learnable weight matrix per layer. Now, to compute $\cW^{\ell}_{k,M}$ using Eq.~\ref{eq:cfw}, we need to know $\cG^{\ell}_{k,i}=\{\cg_{v,k,i}^{\ell}\mid v\in\CV\}$ and $\CZ^{\ell-1}_{k,i}=\{\cz_{v,k,i}^{\ell-1}\mid v\in\CV\}$. From Eq.~\ref{eq:sage}, it is easy to see that for any GraphSAGE model $\Theta_i$, we have:
\begin{align}    
 \mathbf{g}_{v,1,i}^{\ell} = \underbrace{\left(\mathbf{h_{v,i}^{(\ell-1)}} \right)}_{\cz_{v,1,i}^{\ell-1}}\mathbf{W}^{(\ell)}_{1,i}    \\
 \mathbf{g}_{v,2,i}^{\ell} = \underbrace{\left(\mathbf{k_{v,i}^{(\ell-1)}} \right)}_{\cz_{v,2,i}^{\ell-1}}\mathbf{W}^{(\ell)}_{2,i}    
\end{align}

\subsection{Graph Isomorphism Network(GIN)}
The node embedding update equation for \textbf{GIN} is as follows:
\[
\mathbf{h}_v^{(\ell)} = \boldsymbol{\phi}^{(\ell)}\left(\mathbf{h}_v^{(\ell-1)}+\sum_{u \in \mathcal{N}_v} \mathbf{h}_u^{(\ell-1)} \right)
\]
where:
\begin{itemize}
    \item \(\mathbf{h}_v^{(\ell)}\): Updated feature vector of node \(i\) at layer \(\ell\).
    \item \(\mathcal{N}_v\): Set of neighbors of node \(v\).
    \item \(\mathbf{h}_u^{(\ell-1)}\): Feature vector of node \(u\) at layer \(l\).
    \item \(\boldsymbol{\phi}^{(\ell)}\): Trainable \textbf{MLP} at layer \(l\).
\end{itemize}
Here, every node collections messages from its neighbours, as well as itself, and takes the sum of the messages. The term $\left(\mathbf{h}_v^{(\ell-1)}+\sum_{u \in \mathcal{N}_v} \mathbf{h}_u^{(\ell-1)} \right)$ can be computed independent of $\boldsymbol{\phi}^{(\ell)}$ and can be denoted by $\mathbf{k}_v^{(\ell-1)}$. Hence, the node update equation becomes:
\[
\mathbf{h}_v^{\ell} = \boldsymbol{\phi}^{(\ell)}(\mathbf{k}_v^{(\ell-1)})
\]
The node update equation is just an MLP applied on $\mathbf{k}_v^{(\ell-1)}$. A typical $N$ layer MLP is as follows:
\begin{equation}
\label{eq:gin_mlp}
\mathbf{y} = \text{MLP}^{(N)}(\mathbf{x}) = f_N(W_N \cdot f_{N-1}(W_{N-1} \cdot \dots f_1(W_1 \cdot \mathbf{x})))
\end{equation}
\text{Where:} \\
$\mathbf{x}$ is the input vector, \\
$\mathbf{W}_i$ are the weight matrices for each layer, \\
$\mathbf{f}_i$ are the activation functions for each layer, and \\
$\mathbf{y}$ is the output. \\
The operation at layer $\mathbf{n}$ is just a linear transform $\mathbf{W}_n$ followed by an activation function $\mathbf{f}_n$. Hence, the MLP can be broken down into a series of linear transforms. \\
Hence, when applied to the generic framework, $K = N$, i.e, there are $N$ learnable weight matrices per layer. Now, to compute $\cW^{\ell}_{n,M}$ using Eq.~\ref{eq:cfw}, we need to know $\cG^{\ell}_{n,i}=\{\cg_{v,n,i}^{\ell}\mid v\in\CV\}$ and $\CZ^{\ell-1}_{n,i}=\{\cz_{v,n,i}^{\ell-1}\mid v\in\CV\}$. \\
$\cG^{\ell}_{1,i}$ and $\CZ^{\ell-1}_{n,i}$ can simply be obtained by :
\begin{equation}
    \mathbf{g}_{v,1,i}^{\ell} = \underbrace{\left(\mathbf{k_{v,i}^{(\ell-1)}} \right)}_{\cz_{v,1,i}^{\ell-1}}\mathbf{W}^{(\ell)}_{1,i} 
\end{equation}
From eq~\ref{eq:gin_mlp}, we can write $\mathbf{g}_{v,n,i}^{\ell}$ inductively as
\begin{equation}
    \mathbf{g}_{v,n,i}^{\ell} = \underbrace{\left(f_{n-1}\left(\mathbf{g}_{v,n-1,i}^{\ell} \right)\right)}_{\cz_{v,n,i}^{\ell-1}}\mathbf{W}^{(\ell)}_{n,i} 
\end{equation}
to obtain $\cG^{\ell}_{n,i}=\{\cg_{v,n,i}^{\ell}\mid v\in\CV\}$ and $\CZ^{\ell-1}_{n,i}=\{\cz_{v,n,i}^{\ell-1}\mid v\in\CV\}$.
\subsection{Graph Attention Network(GAT)}
The node embedding update equation for \textbf{GAT} before activation is as follows:
\[
\mathbf{h}_v^{(\ell)} = \sigma\left( \sum_{u \in \mathcal{N}_v} \alpha_{uv} \mathbf{h}_u^{(\ell-1)}\mathbf{W}^{(\ell)}  \right)
\]
where:
\begin{itemize}
    \item \(\mathcal{N}_v\): Set of neighbors of node \(v\).
    \item \(\mathbf{h}_u^{(\ell)}\): Feature vector of node \(u\) at layer \(l\).
    \item \(\mathbf{W}^{(\ell)}\): Trainable weight matrix at layer \(l\).
    \item \(\alpha_{vu}\) : Attention coefficients between node $v$ and its neighbour node $u$.
\end{itemize}
The attention coefficients $\alpha_{vu}$ are computed using the attention mechanism. typically involving a self-attention mechanism such as:
\begin{equation}
\alpha_{uv} = \frac{\exp \left( \text{LeakyReLU} \left( \mathbf{a^{(\ell)}}^{\mathtt{T}} [\mathbf{W}^{(\ell)} \mathbf{h}_v^{(\ell-1)} \parallel \mathbf{W}^{(\ell)} \mathbf{h}_u^{(\ell-1)}] \right) \right)}{\sum_{k \in \mathcal{N}_v} \exp \left( \text{LeakyReLU} \left( \mathbf{a^{(\ell)}}^{\mathtt{T}}[\mathbf{W}^{(\ell)} \mathbf{h}_v^{(\ell-1)} \parallel \mathbf{W}^{(\ell)} \mathbf{h}_k^{(\ell-1)}] \right) \right)}
\end{equation}
which involves a learnable vector $\mathbf{a^{(\ell)}}$.\\
Hence, when applied to the generic framework, $K=2$, i.e, there are $2$ learnable weight matrices per layer. \\
For $\mathbf{W}^{(\ell)}_M$, we simply have:
\begin{equation}
    \mathbf{g}_{v,i}^{\ell} = \underbrace{\left( \sum_{u \in \mathcal{N}_v} \alpha_{uv,i}^{\ell} \mathbf{h}_{u,i}^{(\ell-1)} \right)}_{\cz_{v,i}^{\ell-1}}\mathbf{W}^{(\ell)}_{i}  
\end{equation}
where, $\alpha_{uv,i}^{\ell}$ is computed as:
\[
    \alpha_{uv,i}^{\ell} = \frac{\exp \left( \text{LeakyReLU} \left( \mathbf{a_i^{(\ell)}}^{\mathtt{T}} [\mathbf{W}_{i}^{(\ell)} \mathbf{h}_{v,i}^{(\ell-1)} \parallel \mathbf{W}_{i}^{(\ell)} \mathbf{h}_{u,i}^{(\ell-1)}] \right) \right)}{\sum_{k \in \mathcal{N}_v} \exp \left( \text{LeakyReLU} \left( \mathbf{a_{i}^{(\ell)}}^{\mathtt{T}}[\mathbf{W}_{i}^{(\ell)} \mathbf{h}_{v,i}^{(\ell-1)} \parallel \mathbf{W}_{i}^{(\ell)} \mathbf{h}_{k,i}^{(\ell-1)}] \right) \right)}
\]
For $\mathbf{a_M^{(\ell)}}$, we have:
\[
    \mathbf{g}_{uv,i}^{\ell} = \underbrace{\left( [\mathbf{W}_{i}^{(\ell)} \mathbf{h}_{v,i}^{(\ell-1)} \parallel \mathbf{W}_{i}^{(\ell)} \mathbf{h}_{u,i}^{(\ell-1)}] \right)}_{\cz_{uv,i}^{\ell-1}}\mathbf{a}^{(\ell)}_{i}
\]
\subsection{NodeFormer}
NodeFormer follows the general idea of Queries, Keys, and Values present in Transformers. In each transformer layer, we have the $\mathbf{W_{Q}^{\ell}}$, $\mathbf{W_{K}^{\ell}}$ and $\mathbf{W_{V}^{\ell}}$ matrices that are used to compute queries, keys and values for each node as:
\begin{align*}
\mathbf{q_v^{\ell}} &= \mathbf{W_{Q}^{\ell}}\mathbf{z}_v^{\ell-1} \\ 
\mathbf{k_v^{\ell}} &= \mathbf{W_{K}^{\ell}}\mathbf{z}_v^{\ell-1} \\ 
\mathbf{v_v^{\ell}} &= \mathbf{W_{V}^{\ell}}\mathbf{z}_v^{\ell-1} 
\end{align*}
where $\mathbf{z}_v^{\ell-1}$ is the node embedding produced by the previous layer. Additionally, it also has a $\mathbf{W}_{O}^{\ell}$ which is used to aggregate the results of multiple heads to obtain the final node embedding for the layer $\ell$ as follows:
\begin{align*}
\mathbf{z_v^{\ell}} &= \mathbf{W_{O}^{\ell}}\mathbf{z'}_v^{\ell} 
\end{align*}
where $\mathbf{z'}_v^{\ell-1}$ is obtained by applying attention pooling using $\mathbf{q_v^{\ell}}$, $\mathbf{k_v^{\ell}}$ and $\mathbf{v_v^{\ell}}$, according to NodeFormer equation:
\[
    \mathbf{z'}_v^{\ell} = \sum_{u=1}^{|V|} \left(\frac{\kappa\left(\mathbf{q_v^{\ell}},\mathbf{k_u^{\ell}}\right)}{\sum_{w=1}^{|V|} \kappa\left(\mathbf{q_v^{\ell}},\mathbf{k_w^{\ell}}\right)}\right)\mathbf{v_u^{\ell}}
\]
where, $\kappa$ is a kernel measuring pairwise similarity.
All of $\mathbf{W}_{Q,M}^{\ell}$, $\mathbf{W}_{K,M}^{\ell}$, $\mathbf{W}_{V,M}^{\ell}$ and $\mathbf{W}_{O,M}^{\ell}$ can be computed analytically using similar formulation as discussed above for MPNNs.   


\section{Experiments}
\label{app:experiments}
\subsection{Hardware Configuration}
All experiments were conducted on a high-performance computing system with the following specifications:
\begin{itemize}
    \item \textbf{CPU:} 96 logical cores
    \item \textbf{RAM:} 512 GB
    \item \textbf{GPU:} NVIDIA A100-PCIE-40GB
\end{itemize}

\subsection{Software Configuration}
The software environment for our experiments was configured as follows:
\begin{itemize}
    \item \textbf{Operating System:} Linux (Ubuntu 20.04.4 LTS (GNU/Linux 5.4.0-124-generic x86\_64))
    \item \textbf{PyTorch Version:} 1.13.1+cu117
    \item \textbf{CUDA Version:} 11.7
    \item \textbf{PyTorch Geometric Version:} 2.3.1
\end{itemize}

\subsection{Parameters used for \name}
\begin{itemize}
    \item Default number of layers in \gnn: 2, with ReLU in between.
    \item Hidden Dimension: 128
    \item Learning rate: 0.05
    \item Optimizer: Adam
\end{itemize}

\subsection{Baselines}
\label{app:baselines}
We compare our proposed model merging approach against six baselines:

\begin{enumerate}
    \item \textbf{Individual Models}: We train separate GNN models for each task independently without any merging. This serves as an upper bound for task-specific performance.

    \item \textbf{Weight Averaging}: A simple model merging baseline where corresponding parameters of two models are averaged element-wise. While computationally inexpensive, this method often fails when models are misaligned.

    \item \textbf{Git Re-Basin}: A model merging baseline that finds an optimal permutation of one model's parameters to better align with another before averaging. It follows the idea that the models merge better if they are permuted to the same loss basin before averaging.

    \item \textbf{Permute}: Another permutation-based baseline that uses linear sum assignment to find optimal permutation for weight averaging.

    \item \textbf{ZipIt!}: Argues that features of models trained on different tasks may be dissimilar, leading to poor merging using traditional methods. In addition to merging features across both models, it also allows merging within the same model. This allows the combination of features within the same model that are compatible with each other.
    \item \textbf{Surgery, with \wa}: Post-hoc task-specific adapter modules are incorporated on top of the weight-averaged merged model, enhancing performance at the cost of introducing additional task specific parameters
\end{enumerate}
\begin{table*}
\centering
\vspace{-0.1in}
\scalebox{0.8}{
\begin{tabular}{lcccc}
\toprule
\textbf{Dataset} & \textbf{\#Nodes} & \textbf{\#Edges} & \textbf{\#Classes} & \textbf{\#Features} \\
\midrule
Cora~\citep{pmlr-v48-yanga16} & 2,708 & 5,429 & 7 & 1,433 \\
Citeseer~\citep{pmlr-v48-yanga16} & 3,312 & 4,732 & 6 & 3,703\\
Pubmed~\citep{pmlr-v48-yanga16} &19,717 & 44,338 & 3 & 500\\
Arxiv~\citep{NEURIPS2020_fb60d411-ogbn} & 169,443 & 2,315,598 & 40 & 128 \\
WikiCS~\citep{mernyei2020wiki-wikics} & 11,701 & 431,726 & 10 & 300 \\
AmzPhoto~\citep{shchur2018pitfalls} & 7,650 & 238,162 & 8 & 745 \\
AmzComp~\citep{shchur2018pitfalls} & 13,752 & 491,722 & 10 & 767 \\
Reddit~\citep{NIPS2017_5dd9db5e-inductive-rep} & 232,965 & 114,615,892 & 41 & 602 \\
\bottomrule
\end{tabular}}
\vspace{-0.1in}
\caption{\small Datasets used for benchmarking \name.}
\label{tab:datasets1}
\vspace{-0.2in}
\end{table*}

\subsection{Additional Experimental Details}

\begin{itemize}
    \item For the node classification tasks, we used the default train-val-test splits available with the respective datasets.
    \item For link prediction tasks, we generated a 70-10-20 train-val-test split using the RandomLinkSplit function in Pytorch. The ratio of positive to negative links was set to 1.0.
    \item The disjoint label splits were created by taking the nodes that belonged to the first $\frac{N}{2}$ classes in the first dataset, and the nodes belonging to the next $\frac{N}{2}$ classes in the second dataset. $N=$ total different classes in the dataset.
    \item For all the the baselines, the default hyperparameters provided in the source code were used.
\end{itemize}
\begin{table*}[t]
\centering
  \scalebox{0.75}{
\begin{tabular}{ l|cccccccc}
\midrule
\textbf{Dataset} & \textbf{M} & \textbf{Raw} & \textbf{\wa}  & \textbf{\permute} & \textbf{\zipit} & \textbf{\surgery} & \textbf{\name} & \textbf{\name{}++} \\ \midrule
\multirow{4}{*}{\textbf{AmzComp}} & \sage 1 & 95.46 & 58.55 & 84.47 & 66.92 & 85.88 & 93.89 &\cellcolor{Gray}94.11 \\ 
                                & \sage 2 & 92.83 & 75.91  & 69.14 & 74.74 & 69.92 & 91.01 & \cellcolor{Gray}91.51 \\ 
                                \cmidrule{2-9}
                                & \node 1 & 93.33 & 49.15  & - & - & 87.86& \cellcolor{Gray}90.22 & 89.42\\ 
                                & \node 2 & 91.96 & 66.51  & - & - & 80.85 & \cellcolor{Gray}88.85 & 86.71 \\ 
                                \midrule
\multirow{4}{*}{\textbf{WikiCS}} & \sage 1 & 86.64 & 80.36  & 83.35 & 76.00 & 84.14 & 84.09 &\cellcolor{Gray}84.43 \\ 
                                & \sage 2 & 84.99 & 76.80 & 56.85 & 64.50 & 82.23 & \cellcolor{Gray}83.73 &83.56 \\ 
                                \cmidrule{2-9}
                                & \node 1 & 79.21 & 56.12  & - & - & 76.11 & \cellcolor{Gray}79.02 & 76.57 \\ 
                                & \node 2 & 78.54 & 70.19  & - & - & 71.34 & \cellcolor{Gray}75.92 & 72.43\\ 
                                \midrule
\end{tabular}}
\vspace{-0.1in}
\caption{\small\textbf{In-domain Dataset} experiments on \sage and \node. Metric reported: Accuracy$(\%)$. \permute, and \zipit are not applicable for transformer architectures.}
\label{tab:results7}
\end{table*}

\begin{table*}[t]
\centering
  \scalebox{0.75}{
\begin{tabular}{ l|cccccccc}
\midrule
\textbf{Arch.} & \textbf{Datasets} & \textbf{Raw}& \textbf{\wa} & \textbf{\permute} & \textbf{\zipit} & \textbf{\surgery} & \textbf{\name} & \textbf{\name{}++} \\
\midrule
\multirow{4}{*}{\sage} & \textbf{Arxiv} & 74.65 & 55.13 & 58.45 & 59.13 & 68.39 & \cellcolor{Gray}72.85 & 72.78\\
& \textbf{WikiCS} & 78.82 & 72.84 & 56.17 & 56.68 & 69.22 & \cellcolor{Gray}78.65 & 78.63 \\
\cmidrule{2-9}
& \textbf{Arxiv} & 74.65 & 68.77 & 60.00 & 63.60 & 67.85 & 74.24 & \cellcolor{Gray}74.29 \\
& \textbf{Pubmed} & 77.96 & 72.32 & 62.21 & 65.94 & 75.12 & \cellcolor{Gray}78.01 & 77.97 \\
\midrule
\multirow{4}{*}{\node} & \textbf{Cora} & 81.09 & 50.72 & - & - & \cellcolor{Gray}77.61 & 76.59 & 75.27\\
& \textbf{Citeseer} & 81.35 & 71.71  & - & - & \cellcolor{Gray}79.15 & 77.83 & 76.08 \\
\cmidrule{2-9}
& \textbf{Pubmed} & 80.08 & 57.60 & - & - & 75.74 & \cellcolor{Gray}79.78 & 79.10\\
& \textbf{WikiCS} & 74.17 & 62.80 & - & - & 67.48 & \cellcolor{Gray}72.42 & 69.21 \\
\midrule
\end{tabular}}
\vspace{-0.1in}
\caption{\small\textbf{Two different datasets} experiments for \sage and \node. Metric reported: Accuracy$(\%)$. \permute, and \zipit do not support transformer architectures.}
\label{tab:results8}
\vspace{-0.2in}
\end{table*}
\vspace{-0.1in}

\section{Additional Results and Analysis}
\label{sec:additional}
\subsection{Different Datasets Results.}
Tables~\ref{tab:results2},~\ref{tab:results3},~\ref{tab:results4} and~\ref{tab:results5} contain the full results for the \textbf{Different Datasets} experiments, with varying number of models being merged.
\begin{figure}
    \centering
    \includegraphics[scale=0.9]{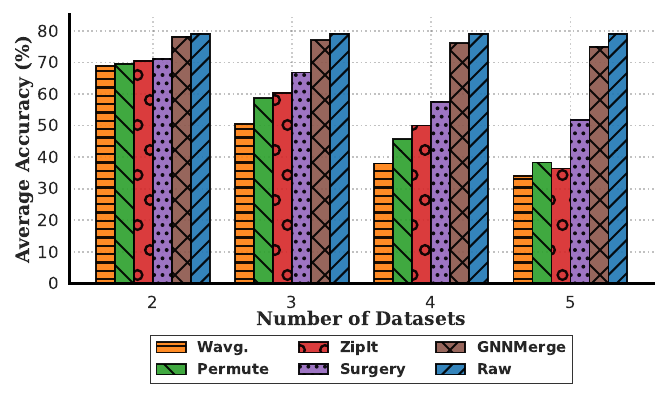}
    \vspace{-10pt} 
    \caption{\small Variation of average accuracy of merging methods as the number of models varies.}
    \label{fig:datasets}
    \vspace{-0.15in}
\end{figure}
\subsection{Generalization to \gnn Architectures} 
\vspace{-0.05in}
We further benchmark \name on \sage and \node in Tables~\ref{tab:results8} and~\ref{tab:results7}. The results follow the same trend observed with \gcn on both architectures, and thereby establishing the robustness of \name to  accommodate diverse \gnn architectures.
\vspace{-0.1in}
\begin{table*}[t]
\centering
  \scalebox{0.8}{
\begin{tabular}{ l|cccccccc}
\toprule
\textbf{Tasks} & \textbf{Raw}& \textbf{Random} & \textbf{\wa} & \textbf{\permute} & \textbf{\zipit} & \textbf{\surgery} & \textbf{\name} & \textbf{\name{}++} \\
\midrule
\textbf{Arxiv-NC}    & 73.10    & 9.31    & 57.99    & 61.03    & 60.72    &  66.54   & \cellcolor{Gray}73.03    & 73.01    \\ 
\textbf{Pubmed-LP}   & 97.05    & 90.22    & 91.32    & 91.78    & 91.50    &  93.67   & 96.16    & \cellcolor{Gray}96.23    \\ \midrule
\textbf{WikiCS-NC}   & 79.32    & 13.99    & 75.18    & 74.79    & 75.88    &  77.02   & 78.79    & \cellcolor{Gray}78.88    \\ 
\textbf{Cora-LP}     & 94.34    & 83.62    & 77.28     & 76.90    & 80.56    &  88.39   & 94.12    & \cellcolor{Gray}94.45    \\ \midrule
\textbf{WikiCS-NC}   & 79.32    & 22.87   & 69.46    & 71.30   & 73.35    &  76.93   & 78.88   & \cellcolor{Gray}78.91    \\ 
\textbf{Pubmed-LP}   & 97.05    & 91.50   & 88.70     & 90.07   & 89.45    &  92.71   & 96.26   & \cellcolor{Gray}96.37    \\ \midrule
\end{tabular}}
\vspace{-0.1in}
\caption{\small\textbf{Different Tasks.} \name and \name{}++ compared with baselines on merging two models trained for different tasks on different datasets. NC: Node Classification, metric reported: Accuracy$(\%)$. LP: Link Prediction, metric reported: ROC-AUC.}
\label{tab:results6}
\vspace{-0.2in}
\end{table*}\\
\subsection{Different Tasks} Table~\ref{tab:results6} presents results for a more challenging scenario: merging \gcn{}s trained on two distinct tasks—node classification and link prediction—across two different datasets. In most cases, baseline methods perform no better than a randomly initialized \gcn on the link prediction task. Preserving node classification accuracy leads to a severe degradation in link prediction AUC for the baselines. In contrast, \name and \name{}++ achieve accuracies comparable to the individual models on their respective tasks. 
\subsection{Speed Efficiency} 
\label{sec:time}
\vspace{-0.05in}
\begin{wraptable}{r}{0.4\textwidth}
\centering
\vspace{-0.1in}
\resizebox{\linewidth}{!}{
\begin{tabular}{lccc}
\toprule
\textbf{Dataset} & \textbf{Scratch Train Time} & \textbf{\name} & \textbf{\name{}++} \\
\midrule
Arxiv & 24.19s  & 3.75s  & 1.67s  \\
Reddit  & 697.88s  & 102.99s  & 5.12s  \\
\bottomrule
\end{tabular}
}
\vspace{-0.1in}
\caption{\small Running times for the \textbf{In-domain dataset} task.}
\vspace{-0.2in}
\label{tab:efficiency}
\end{wraptable}
Table~\ref{tab:efficiency} compares the time for training a \gcn from scratch vs. merging two pre-trained models on disjoint label splits using \name and \name{}++. \name provides a \textbf{7×} speedup on Reddit, while \name{}++, benefiting from its analytical solution, enables instantaneous merging with a remarkable \textbf{136×} speedup. Furthermore, \name{}++ only uses CPU. This result highlights the significant potential of model merging for \gnns.

\vspace{-0.05in}
\subsection{Data Efficiency}
\subsubsection{Target Node Sampling}
\begin{figure*}[]
    \centering
    \subfloat[Normalized Test Accuracy on Arxiv as the target nodes sampling ratio is varied]{
    \label{fig:sampling1}
    \includegraphics[width=0.4\textwidth]{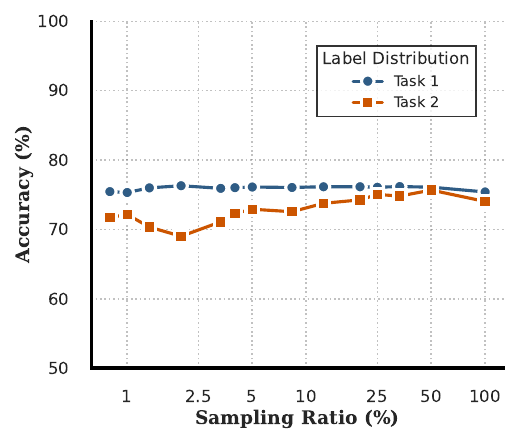}}
    \hspace{1cm}
    \subfloat[Normalized Test Accuracy on Reddit as the target nodes sampling ratio is varied]{
    \label{fig:sampling2}
    \includegraphics[width=0.4\textwidth]{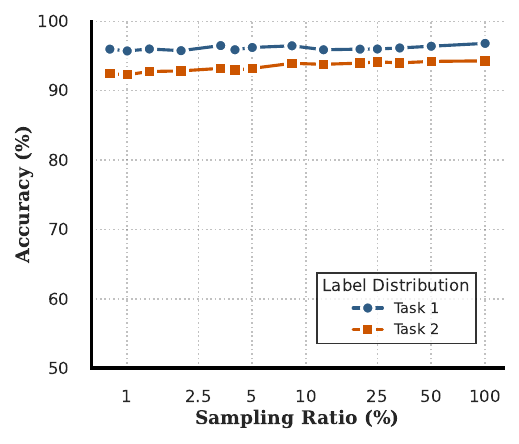}}
\end{figure*}

The objective function described in the main paper is designed to align the embeddings of all nodes within a task's dataset. However, aligning only a subset of nodes may suffice to achieve a comparable alignment quality for the entire dataset, as the information encoded in the embeddings of a representative subset can effectively propagate to the remaining nodes through the graph structure. So the question is: \textit{how many nodes do you we need to get a good alignment?} \\
Figure \ref{fig:sampling1} and \ref{fig:sampling2} depict the variation in test accuracies as the percentage of nodes utilized for alignment is varied on the arxiv and reddit datasets.\\
For the arxiv dataset, there is a small drop in accuracy only as the sampling ratio reaches about $2.5\%$. For the reddit dataset, even a sampling ratio as small as $0.8\%$ has no practical effect on the model merging performance. This suggests that the method can be accelerated by aligning a smaller subset of nodes without compromising effectiveness.

\subsubsection{1-Hop Neighbour Condensation}
\begin{figure*}[h!]
    \centering
    \subfloat[Graph Size(Edges) of Arxiv as the sampling ratio is varied]{
    \label{fig:sampling3}
    \includegraphics[width=0.36\textwidth]{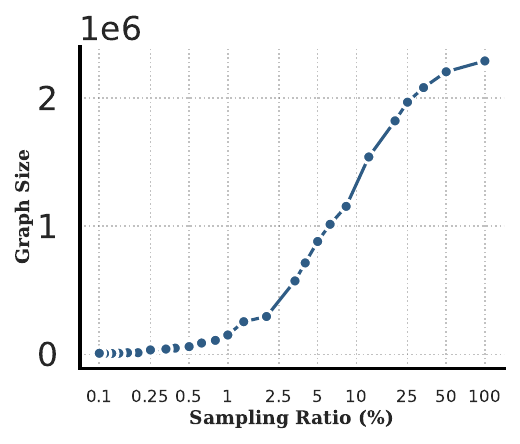}}
    \hspace{1cm}
    \subfloat[Graph Size(Edges) of Reddit as the sampling ratio is varied]{
    \label{fig:sampling4}
    \includegraphics[width=0.36\textwidth]{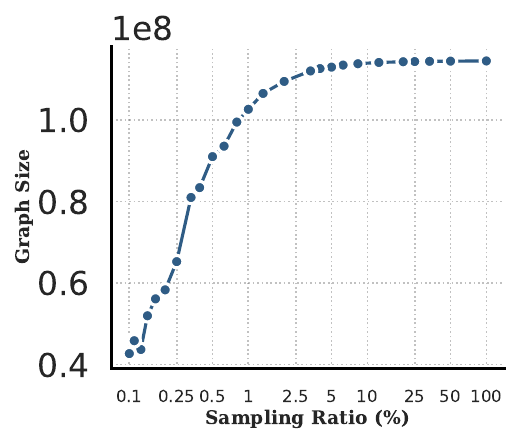}}
\end{figure*}

In an \mpnn architecture, each layer computes node embeddings by aggregating messages from a node's 1-hop neighbors. As a result, at any given layer, a target node's embedding depends exclusively on its immediate neighbors, if the input is fixed. This property enables the graph to be reduced to only the target nodes and their 1-hop neighbors, significantly decreasing its size. When combined with node sampling, this leads to a substantial reduction in both memory requirements and merging time. Figures \ref{fig:sampling3} and \ref{fig:sampling4} illustrate the variation in memory consumption across different node sampling levels for the Arxiv and Reddit datasets, respectively. The overall sampling procedure leads to 3 benefits:
\begin{enumerate}
    \item \textbf{Reduced Memory Requirement:} The required graph size after 1-hop neighbor condensation dramatically falls as the sampling ratio is reduced.
    \item \textbf{Reduced Convergence Time:} As the model aligns a smaller subset of nodes, the complexity of the loss function is reduced, resulting in faster convergence.
    \item \textbf{Reduced Forward Pass Time:} Forward pass time for \gnn architecture is $O(E)$. Reduction in edges leads to faster forward pass.
\end{enumerate}

\newpage
\section{Additional Visualisation}
\label{app:visual}
\begin{figure*}[h!]
    \centering
    \vspace{-0.1in}
    \subfloat[]{
    \label{fig:emb4}
    \includegraphics[width=0.32\textwidth]{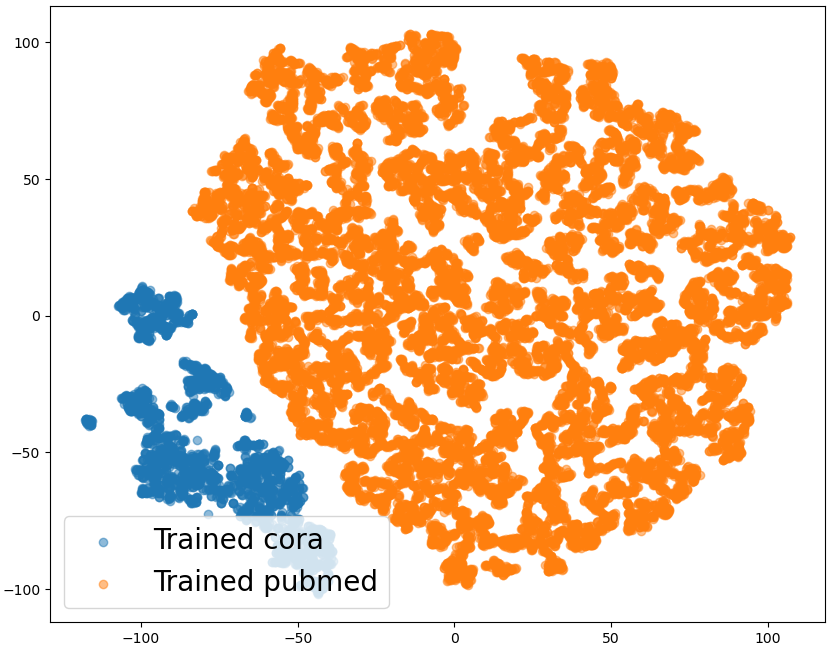}}
    \subfloat[]{
    \label{fig:emb5}
    \includegraphics[width=0.3\textwidth]{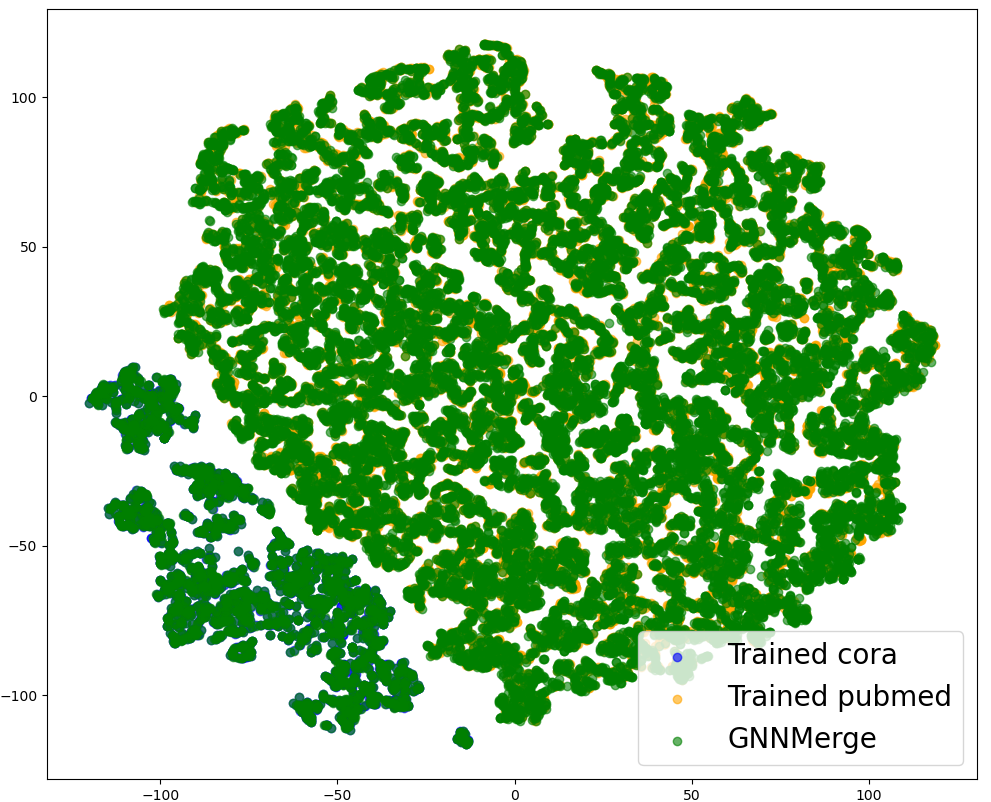}}
    \subfloat[]{
    \label{fig:emb6}
    \includegraphics[width=0.3\textwidth]{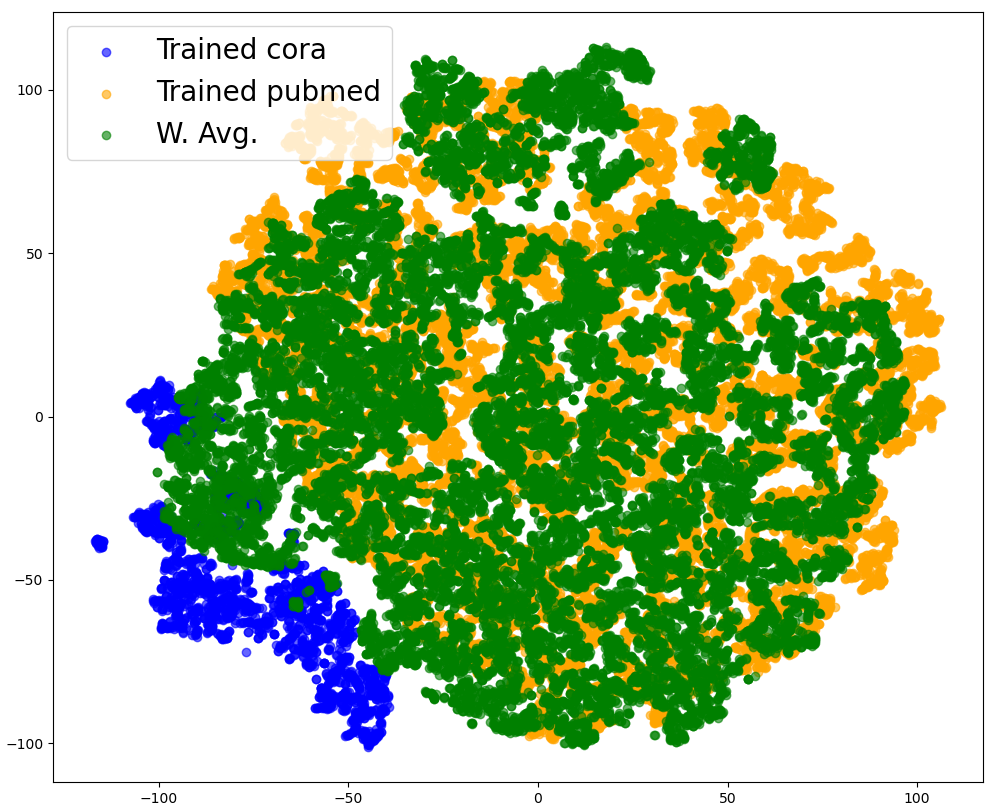}}
    \vspace{-6pt}
    \caption{\small Visual Illustration of embedding alignment using \name{} and \wa as the merging methods.}
    \label{fig:cora_pubmed}
\end{figure*}

\begin{figure*}[h!]
    \centering
    \vspace{-0.3in}
    \subfloat[]{
    \label{fig:emb7}
    \includegraphics[width=0.32\textwidth]{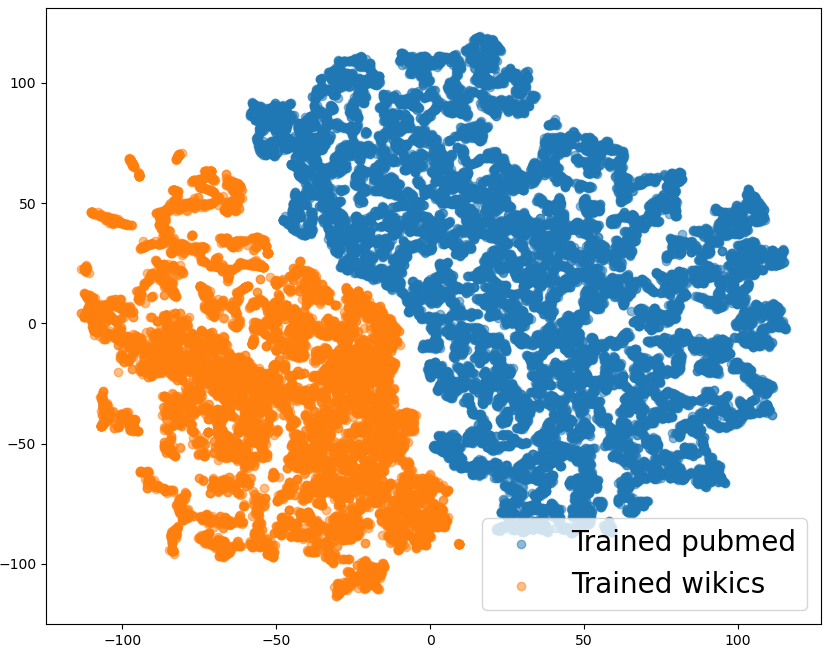}}
    \subfloat[]{
    \label{fig:emb8}
    \includegraphics[width=0.3\textwidth]{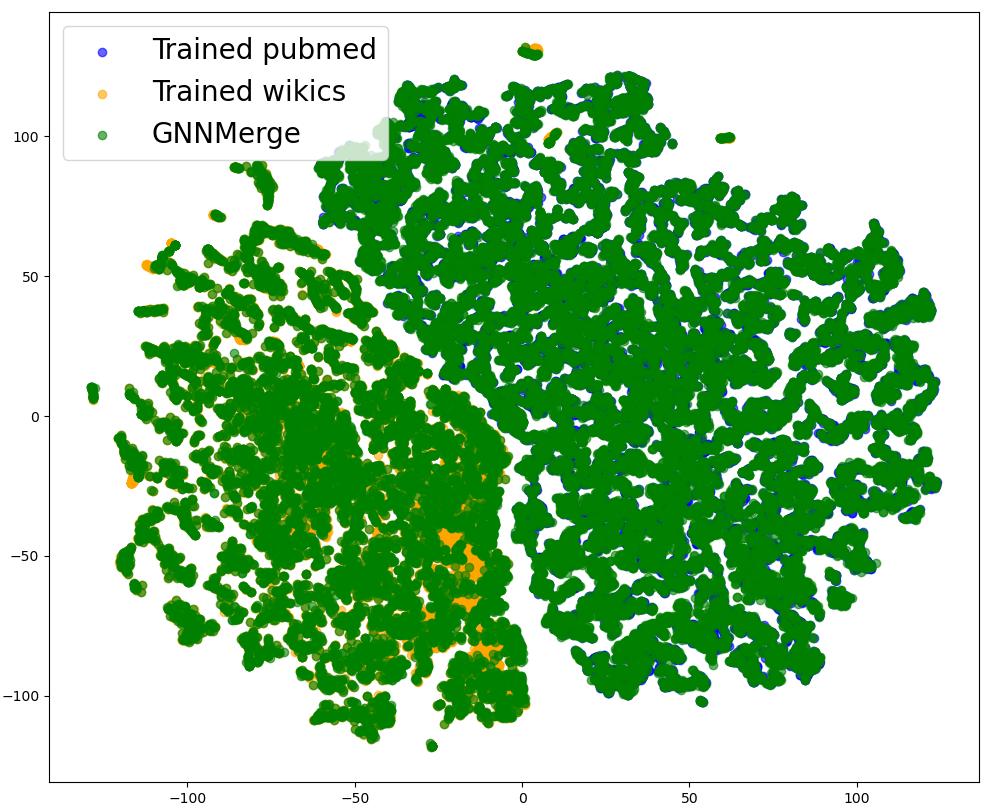}}
    \subfloat[]{
    \label{fig:emb9}
    \includegraphics[width=0.3\textwidth]{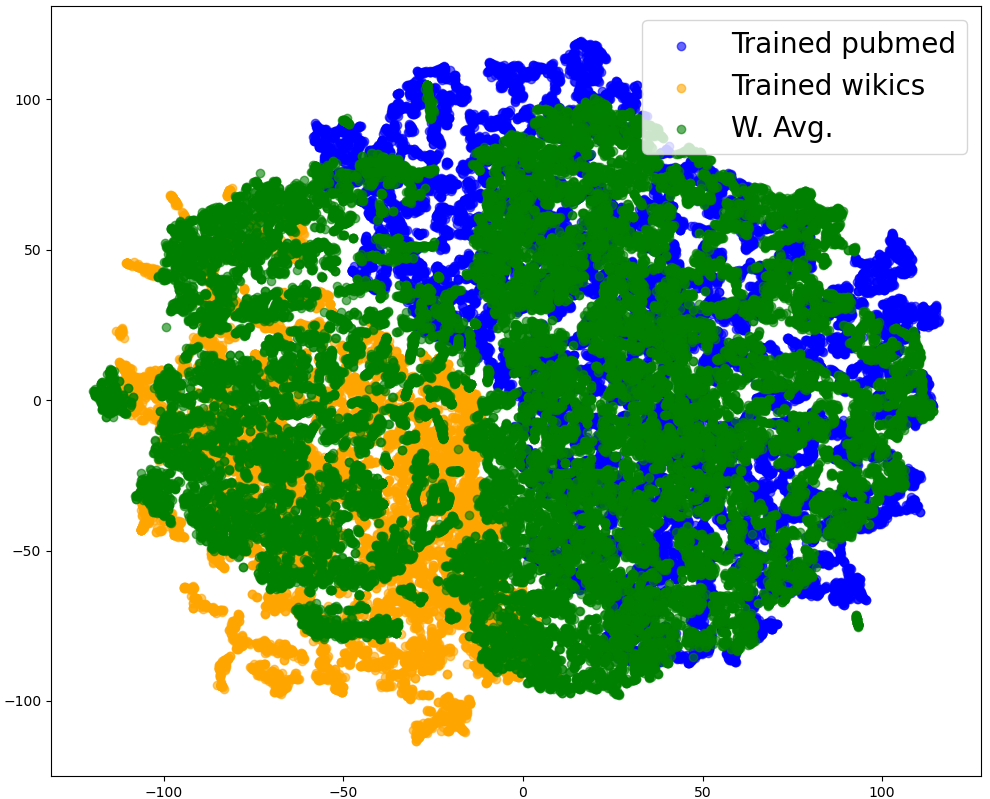}}
    \vspace{-6pt}
    \caption{\small Visual Illustration of embedding alignment using \name{} and \wa as the merging methods.}
    \label{fig:pubmed_wikics}
\end{figure*}

Following the discussion in section~\ref{sec:visual}, we present additional node embedding plots in this section. For Cora+Pubmed(fig~\ref{fig:cora_pubmed}) and Pubmed+WikiCS(fig~\ref{fig:pubmed_wikics}), similar type of behaviour as discussed in section~\ref{sec:visual} is observed. \name manages to completely overlap the embeddings produced by the base models, leading to good performance of the merged model.

\begin{figure*}[h!]
    \centering
    \vspace{-0.1in}
    \subfloat[]{
    \label{fig:emb10}
    \includegraphics[width=0.32\textwidth]{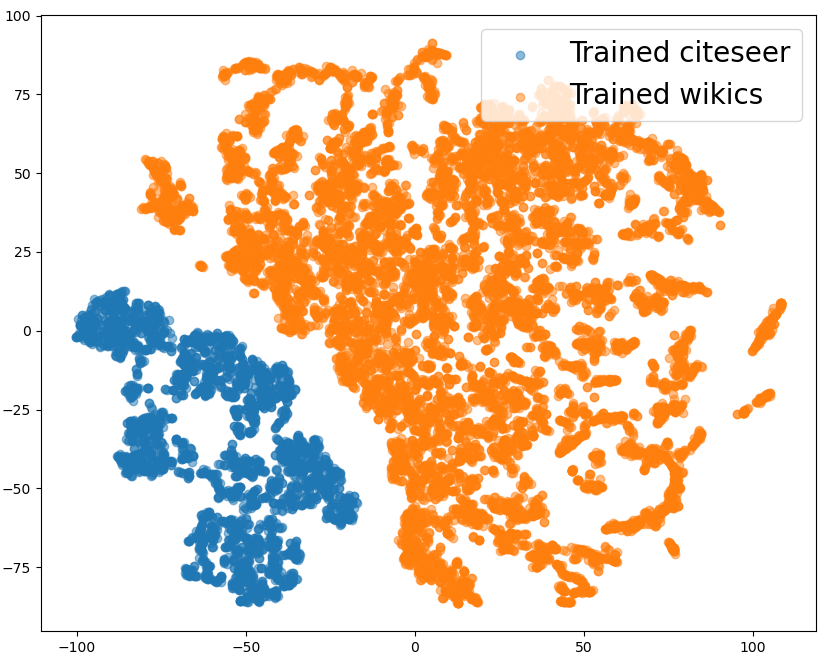}}
    \subfloat[]{
    \label{fig:emb11}
    \includegraphics[width=0.3\textwidth]{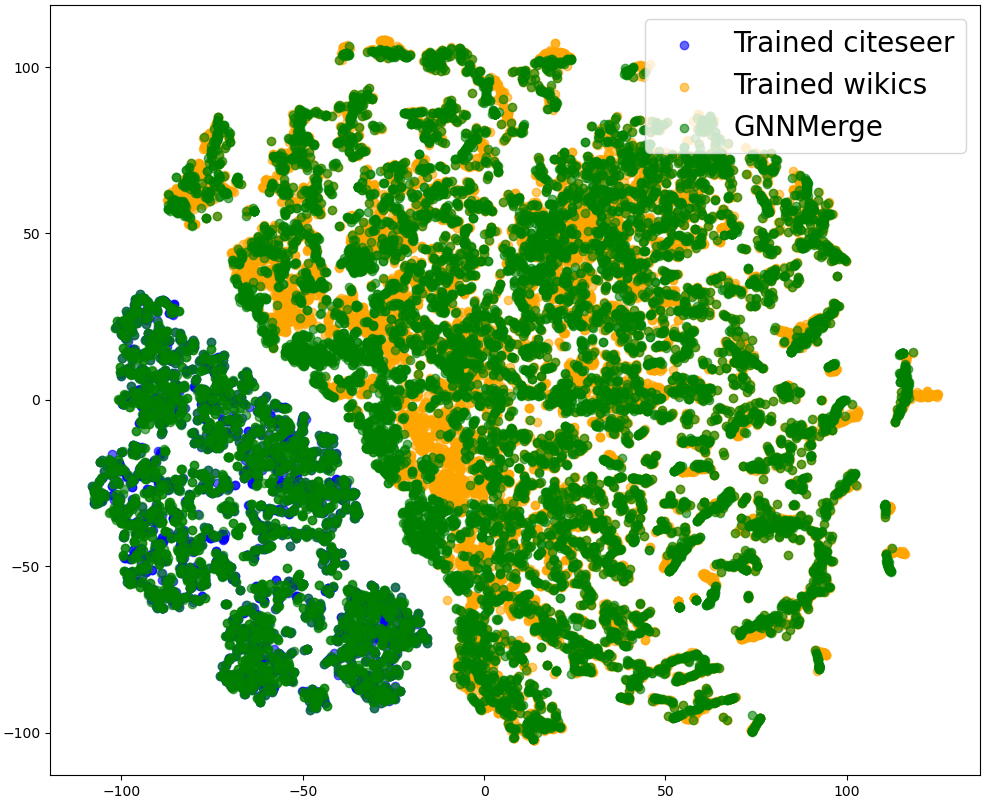}}
    \subfloat[]{
    \label{fig:emb12}
    \includegraphics[width=0.3\textwidth]{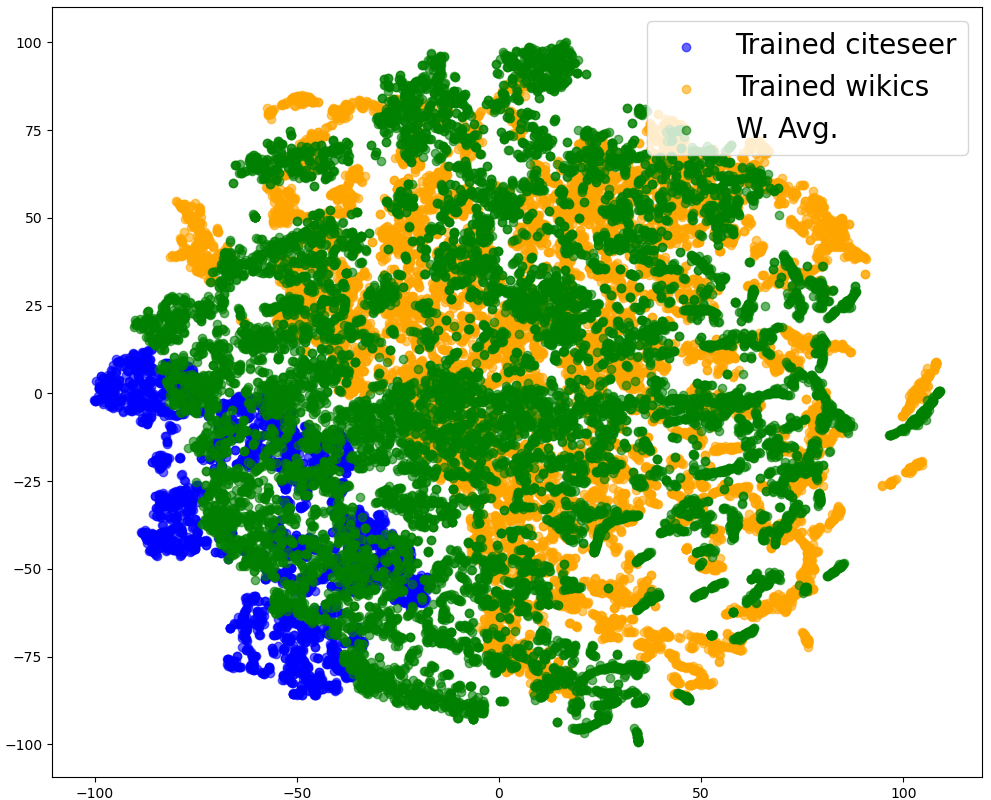}}
    \vspace{-6pt}
    \caption{\small Visual Illustration of embedding alignment using \name{} and \wa as the merging methods.}
    \label{fig:citeseer_wikics}
\end{figure*}

We also present the plots for Citeseer+Wikics(fig~\ref{fig:citeseer_wikics}). Notably, in table~\ref{tab:results2half}, \name{}++ suffers a $1.1\%$ accuracy drop on WikiCS. Compared to an average drop of $0.26\%$, this makes Citeseer+Wikics one of the difficult cases. This difficulty is actually highlighted by the fact that the overlap in fig~\ref{fig:emb11} isn't as good as the other cases for \name. This actually depicts the importance of a good alignment and why our method works well if a good alignment is possible.
\begin{table*}[h!]
\centering
\scalebox{0.85}{
\begin{tabular}{l|cccccccc}
\toprule
\textbf{Datasets} & \textbf{Raw}& \textbf{\wa} & \textbf{\git} & \textbf{\permute} & \textbf{\zipit} & \textbf{\surgery} & \textbf{\name} & \textbf{\name{}++} \\
\midrule
\textbf{Cora}    & 81.86    & 74.74    & 77.80    & 75.43    & 72.72    & 70.44    & \cellcolor{Gray}82.01    & 81.76    \\ 
\textbf{Citeseer} & 81.97   & 70.88    & 25.39    & 73.98    & 77.58    & \cellcolor{Gray}79.81    & 78.52    & 78.52    \\ \midrule
\textbf{Cora}    & 81.86    & 74.54    & 77.12    & 75.19    & 76.16    & 72.66    & \cellcolor{Gray}81.96    & 81.76    \\ 
\textbf{Arxiv}   & 73.10    & 47.05    & 6.11     & 46.25    & 50.16    & 55.79    & 67.07    & \cellcolor{Gray}67.13    \\ \midrule
\textbf{Cora}    & 81.86    & 79.09    & 78.19    & 80.46    & \cellcolor{Gray}81.43    & 75.78    & \cellcolor{Gray}81.43    & 81.14    \\ 
\textbf{Pubmed}  & 79.02    & 76.45    & 51.22    & 77.66    & 78.21    & 75.05    & \cellcolor{Gray}79.17    & 79.08    \\ \midrule
\textbf{Cora}    & 81.86    & 70.87    & 78.62    & 74.56    & 76.54    & 76.49    & 81.57    & \cellcolor{Gray}81.62    \\ 
\textbf{WikiCS}  & 79.32    & 65.79    & 25.08    & 68.10    & 68.70    & 68.84    & 78.65    & \cellcolor{Gray}79.04    \\ \midrule
\textbf{Citeseer} & 81.97   & 73.39    & 78.36    & 76.33    & 77.74    & \cellcolor{Gray}81.81    & 81.03    & 81.34    \\ 
\textbf{Arxiv}   & 73.10    & 44.84    & 5.81     & 53.04    & 53.70    & 52.25    & 67.37    & \cellcolor{Gray}67.48    \\ \midrule
\textbf{Citeseer} & 81.97   & 78.09    & 80.25    & 79.15    & 78.68    & 79.50    & \cellcolor{Gray}82.91    & 82.44    \\ 
\textbf{Pubmed}  & 79.02    & 75.94    & 22.23    & 78.47    & 77.25    & 68.69    & \cellcolor{Gray}79.14    & 79.04    \\ \midrule
\textbf{Citeseer} & 81.97   & 67.54    & 71.78    & 73.19    & 74.92    & 79.56    & 82.44    & \cellcolor{Gray}82.60    \\ 
\textbf{WikiCS}  & 79.32    & 60.27    & 22.90    & 61.99    & 63.28    & 71.19    & 78.00    & \cellcolor{Gray}78.21    \\ \midrule
\textbf{Arxiv}   & 73.10    & 68.43    & 53.12    & 53.56    & 50.11    & 60.46    & \cellcolor{Gray}72.21    & 71.98    \\ 
\textbf{WikiCS}  & 79.32    & 66.89    & 25.98    & 61.55    & 67.16    & 72.40    & \cellcolor{Gray}79.01    & 78.67    \\ \midrule
\textbf{Arxiv}   & 73.10    & 61.4     & 60.47    & 57.64    & 59.05    & 57.66    & 72.62    & \cellcolor{Gray}72.65    \\ 
\textbf{Pubmed}  & 79.02    & 74.28    & 20.88    & 78.04    & 78.12    & 75.39    & 79.08    & \cellcolor{Gray}79.13    \\ \midrule
\textbf{Pubmed}  & 79.02    & 76.20    & 67.88    & 75.81    & 75.16    & 74.97    & \cellcolor{Gray}78.96    & \cellcolor{Gray}78.96    \\ 
\textbf{WikiCS}  & 79.32    & 70.68    & 8.02     & 69.95    & 73.16    & 69.36    & \cellcolor{Gray}79.39    & 78.89    \\ \midrule
\textbf{Average} & 79.05 & 68.86 & 46.86 & 69.51 & 70.49 & 71.10 & \cellcolor{Gray}\textbf{78.12} & \textbf{78.07}\\ \bottomrule
\end{tabular}}
\caption{\small\textbf{Two Different Datasets Results.} \name and \name{}++ compared with baselines on merging $2$ models trained on $2$ distinct datasets. Metric reported: Accuracy $(\%)$}
\label{tab:results2}
\end{table*}

\begin{table*}[h!]
\centering
\begin{tabular}{ l|cccccccc}
\toprule
\textbf{Datasets} & \textbf{Raw}& \textbf{\wa} & \textbf{\permute} & \textbf{\zipit} & \textbf{\surgery} & \textbf{\name} & \textbf{\name{}++} \\
\midrule
\textbf{Citeseer}    & 81.97    & 50.15      & 67.71    & 65.20    & 77.85    & 78.05    & \cellcolor{Gray}78.21    \\ 
\textbf{Cora}        & 81.86   & 58.65      & 61.60    & 63.39    & 73.20   & \cellcolor{Gray}82.10    & 81.96    \\ 
\textbf{Arxiv}        & 73.10   & 31.68      & 32.94    & 42.22    & 55.32    & 62.05    & \cellcolor{Gray}62.42    \\ \midrule
\textbf{Citeseer}    & 81.97    & 61.75      & 67.86    & 59.24    & \cellcolor{Gray}80.40    & 78.05    & 77.27    \\ 
\textbf{Cora}        & 81.86   & 66.24      & 67.94    & 67.45    & 72.02    & 81.81    & \cellcolor{Gray}81.82    \\ 
\textbf{Pubmed}        & 79.02   & 78.57      & 78.80    & 76.71    & 71.46    & 79.12    & \cellcolor{Gray}79.15    \\ \midrule
\textbf{Citeseer}    & 81.97    & 42.47      & 63.79    & 67.55    & \cellcolor{Gray}80.72    & 78.52    & 78.21    \\ 
\textbf{Cora}        & 81.86   & 52.75      & 48.98    & 74.03    & 75.98    & 81.81    & \cellcolor{Gray}81.82    \\ 
\textbf{WikiCS}        & 79.32   & 33.82      & 43.37    & 34.39    & 58.79    & 77.49    & \cellcolor{Gray}77.73    \\ \midrule
\textbf{Citeseer}    & 81.97    & 45.92      & 61.44    & 59.71    & 80.56    & 82.13    & \cellcolor{Gray}82.29    \\ 
\textbf{Arxiv}        & 73.10   & 10.31      & 35.74    & 37.95    & 53.25    & 66.15    & \cellcolor{Gray}66.39    \\ 
\textbf{WikiCS}        & 79.32   & 26.50      & 35.57    & 30.13    & 56.16    & 77.27    & \cellcolor{Gray}77.46    \\ \midrule
\textbf{Citeseer}    & 81.97    & 63.32      & 68.96    & 64.42    & 75.92    & \cellcolor{Gray}81.66    & 81.35    \\ 
\textbf{Pubmed}        & 79.02   & 74.40      & 76.10    & 75.14    & 74.69    & 79.18   & \cellcolor{Gray}79.29    \\ 
\textbf{Arxiv}        & 73.10   & 27.44      & 48.33    & 47.35    & 51.51    & 66.19    & \cellcolor{Gray}66.38    \\ \midrule
\textbf{Citeseer}    & 81.97    & 50.47      & 69.12    & 69.59    & 76.37    & \cellcolor{Gray}82.60    & 81.82    \\ 
\textbf{Pubmed}        & 79.02   & 75.71      & 73.80    & 76.92    & 73.70    & 79.05    & \cellcolor{Gray}79.09    \\ 
\textbf{WikiCS}        & 79.32   & 46.26      & 48.99    & 63.43    & 48.58    & \cellcolor{Gray}77.61    & 77.53    \\ \midrule
\textbf{Cora}         & 81.86    & 42.40      & 61.17    & 64.55    & 72.61    & 81.76    & \cellcolor{Gray}81.87    \\ 
\textbf{Arxiv}        & 73.10   & 18.18      & 37.15    & 44.35    &  40.82   & 65.75    & \cellcolor{Gray}65.80    \\ 
\textbf{WikiCS}        & 79.32   & 45.21      & 58.09    & 62.15    & 58.95    & \cellcolor{Gray}78.19    & 78.13    \\ \midrule
\textbf{Cora}         & 81.86    & 58.51      & 66.44    & 64.36    & 74.17    & \cellcolor{Gray}81.62    & 81.53    \\ 
\textbf{Pubmed}        & 79.02   & 74.85      & 71.63    & 78.11    & 73.75    & 79.17    & \cellcolor{Gray}79.24    \\ 
\textbf{Arxiv}        & 73.10   & 32.56      & 39.09    & 44.06    & 47.80    & \cellcolor{Gray}66.04    & 66.03    \\ \midrule
\textbf{Cora}         & 81.86    & 59.62      & 71.27    & 67.89    & 72.25    & \cellcolor{Gray}81.14    & 81.00    \\ 
\textbf{Pubmed}        & 79.02   & 78.69      & 76.75    & 77.34    & 74.85    & \cellcolor{Gray}79.10    & 79.01    \\ 
\textbf{WikiCS}        & 79.32   & 52.35      & 59.91    & 62.57    & 68.40    & \cellcolor{Gray}78.68    & 78.60    \\ \midrule
\textbf{Pubmed}       & 79.02    & 76.23      & 77.90    & 75.07    & 73.01    & 79.10    & \cellcolor{Gray}79.13    \\ 
\textbf{Arxiv}        & 73.10   & 27.84      & 38.69    & 41.12    & 51.57    & \cellcolor{Gray}71.47    & 71.41    \\ 
\textbf{WikiCS}        & 79.32   & 52.83      & 51.25    & 54.66    & 59.41    & 78.10    & \cellcolor{Gray}78.18 \\ \midrule
\textbf{Average} & 79.05 & 50.52 & 58.68 & 60.36 & 66.80 & \cellcolor{Gray}\textbf{77.03} & \textbf{77.00}\\ \midrule
\end{tabular}
\caption{\small\textbf{Three Different Datasets Results.} \name and \name{}++ compared with baselines on merging $3$ models trained on $3$ distinct datasets. Metric reported: Accuracy $(\%)$}
\label{tab:results3}
\end{table*}

\begin{table*}[h!]
\centering
\begin{tabular}{ l|cccccccc}
\toprule
\textbf{Datasets} & \textbf{Raw}& \textbf{\wa} & \textbf{\permute} & \textbf{\zipit} & \textbf{\surgery} & \textbf{\name} & \textbf{\name{}++} \\
\midrule
\textbf{Citeseer}    & 81.97    & 36.67      & 50.47    & 52.03    & 76.66    & \cellcolor{Gray}78.05    & 77.43   \\ 
\textbf{Cora}        & 81.86   & 35.10      & 41.58    & 44.39    & 61.15   & \cellcolor{Gray}82.20    & 81.82    \\ 
\textbf{Arxiv}        & 73.10   & 05.93      & 13.36    & 14.20    & 42.13    & 61.57    & \cellcolor{Gray}62.21    \\ 
\textbf{WikiCS}    & 81.97    & 23.70      & 31.45    & 32.59    & 57.89   & 77.01    & \cellcolor{Gray}77.03    \\ \midrule
\textbf{Citeseer}    & 81.97    & 36.52      & 60.18    & 53.44    & 72.57    & \cellcolor{Gray}77.89    & 77.74    \\ 
\textbf{Cora}        & 81.86   & 48.83      & 44.48    & 54.15    & 45.51   & \cellcolor{Gray}82.44    & 82.16    \\ 
\textbf{Pubmed}        & 73.10   & 73.76      & 74.20    & 75.87    & 74.32    & 79.30    & \cellcolor{Gray}79.36    \\ 
\textbf{Arxiv}    & 81.97    & 10.47      & 15.64    & 21.36    & 46.22    & \cellcolor{Gray}61.77    & 61.32    \\ \midrule
\textbf{Citeseer}    & 81.97    & 37.93      & 53.76    & 59.87    & 77.18    & \cellcolor{Gray}77.89    & 77.74    \\ 
\textbf{Cora}        & 81.86   & 41.53      & 46.27    & 57.73    & 73.74   & \cellcolor{Gray}81.81    & 81.67    \\ 
\textbf{Pubmed}        & 73.10   & 77.43      & 73.96    & 75.53    & 73.02    & 79.01    & \cellcolor{Gray}79.07    \\ 
\textbf{WikiCS}    & 81.97    & 34.51      & 37.54    & 39.45    & 52.32    & 77.03    & \cellcolor{Gray}77.10    \\ \midrule
\textbf{Citeseer}    & 81.97    & 38.24      & 54.07    & 60.03    & 73.47    & \cellcolor{Gray}82.28    & 81.97    \\ 
\textbf{Pubmed}        & 81.86   & 73.31      & 75.62    & 69.18    & 72.84   & 79.11    & \cellcolor{Gray}79.13    \\ 
\textbf{Arxiv}        & 73.10   & 05.92      & 22.20    & 32.63    & 43.08    & 65.15   & \cellcolor{Gray}65.30    \\ 
\textbf{WikiCS}    & 81.97    & 28.03     & 33.74    & 34.22    & 40.59    & 76.58    & \cellcolor{Gray}76.62    \\ \midrule
\textbf{Cora}    & 81.97    & 31.38      & 51.35    & 55.31    & 71.00    & \cellcolor{Gray}81.72    & 81.67    \\ 
\textbf{Pubmed}  & 81.86   & 74.18      & 66.79    & 77.44    & 73.79   & 79.11    & \cellcolor{Gray}79.18    \\ 
\textbf{Arxiv}   & 73.10   & 06.96      & 23.62    & 34.99    & 39.24    & \cellcolor{Gray}64.93    & 64.59    \\ 
\textbf{WikiCS}   & 81.97    & 39.78      & 42.73    & 55.27    & 57.99    & \cellcolor{Gray}77.71    & 77.42    \\ \midrule

\textbf{Average} & 79.05 & 38.01 & 45.65 & 49.98 & 57.54 & \cellcolor{Gray}\textbf{76.12} & \textbf{76.03} \\ \midrule
\end{tabular}
\caption{\small\textbf{Four Different Datasets Results.} \name and \name{}++ compared with baselines on merging $4$ models trained on $4$ distinct datasets. Metric reported: Accuracy $(\%)$}
\label{tab:results4}
\end{table*}

\begin{table*}[h!]
\centering
\begin{tabular}{ l|cccccccc}
\toprule
\textbf{Datasets} & \textbf{Raw}& \textbf{\wa} & \textbf{\permute} & \textbf{\zipit} & \textbf{\surgery} & \textbf{\name} & \textbf{\name{}++} \\
\midrule
\textbf{Citeseer}    & 81.97    & 37.46      & 46.39    & 38.55    & 45.63    & 77.12    & \cellcolor{Gray}77.59    \\ 
\textbf{Cora}        & 81.86   & 29.49      & 32.59    & 31.72    & 63.66   & \cellcolor{Gray}81.72    & 81.58    \\ 
\textbf{Pubmed}        & 79.02   & 72.92      & 69.45    & 76.92    & 61.41    & 79.07    & \cellcolor{Gray}79.27    \\ 
\textbf{Arxiv}    & 73.10    & 05.87      & 07.97    & 06.05    & 40.18    &\cellcolor{Gray} 60.27    & \cellcolor{Gray}60.27    \\ 
\textbf{WikiCS}    & 79.32    & 24.54      & 34.85    & 28.57    & 46.40    & \cellcolor{Gray}76.24    & 76.19    \\ \midrule
\textbf{Average} & 79.05 & 34.05 & 38.25 & 36.36 & 51.45 & \textbf{74.88}& \cellcolor{Gray}\textbf{74.98} \\ \midrule
\end{tabular}
\caption{\small\textbf{Five Different Datasets Results.} \name and \name{}++ compared with baselines on merging $5$ models trained on $5$ distinct datasets. Metric reported: Accuracy $(\%)$}
\label{tab:results5}
\end{table*}

\end{document}